\documentclass[runningheads]{llncs}

\usepackage{eccv}

\usepackage{eccvabbrv}

\usepackage[table]{xcolor} 
\usepackage{adjustbox}     
\newcommand{\hl}[1]{\cellcolor{gray!15}{#1}} 
\usepackage{graphicx}
\usepackage{graphicx}
\usepackage{enumitem}
\usepackage{booktabs}
\usepackage{algorithm}
\usepackage{algorithmic}
\usepackage{booktabs}
\usepackage{multirow}
\usepackage{graphicx}
\usepackage[table]{xcolor}
\usepackage{booktabs,multirow}
\usepackage[table]{xcolor}
\usepackage{longtable}
\usepackage{booktabs}
\usepackage{booktabs}
\usepackage{multirow}
\usepackage{colortbl}
\usepackage{xcolor}
\usepackage{graphicx}
\usepackage{algorithm}
\usepackage{algorithmic}
\usepackage{bm}
\usepackage{tcolorbox}
\usepackage{enumitem}
\tcbuselibrary{skins, breakable}
\usepackage{pifont}
\usepackage{xcolor}
\usepackage{graphicx}
\usepackage{xcolor, graphicx, booktabs, tabularx}
\definecolor{halred}{RGB}{200,30,30}
\definecolor{recgreen}{RGB}{0,130,70}
\definecolor{corblue}{RGB}{30,80,180}
\usepackage[accsupp]{axessibility}  

\usepackage{kotex}

\usepackage{hyperref}

\usepackage{orcidlink}

\begin{document}

\title{ReflectCAP: Detailed Image Captioning with Reflective Memory}



\author{Kyungmin Min\inst{1}, Minbeom Kim\inst{1}, Kang-il Lee\inst{2}, \\
Seunghyun Yoon\inst{3}, Kyomin Jung\inst{1,2}\thanks{Corresponding authors}
}

\authorrunning{Min et al.}

\institute{IPAI, Seoul National University \and
Dept. of ECE, Seoul National University \and Adobe Research \\
\email{\{kyungmin97, kjung\}@snu.ac.kr}}

\maketitle

\begin{abstract}
Detailed image captioning demands both factual grounding and fine-grained coverage, yet existing methods have struggled to achieve them simultaneously. We address this tension with Reflective Note-Guided Captioning (ReflectCAP), where a multi-agent pipeline analyzes what the target large vision-language model (LVLM) consistently hallucinates and what it systematically overlooks, distilling these patterns into reusable guidelines called Structured Reflection Notes. At inference time, these notes steer the captioning model along both axes—what to avoid and what to attend to—yielding detailed captions that jointly improve factuality and coverage. Applying this method to 8 LVLMs spanning the GPT-4.1 family, Qwen series, and InternVL variants, ReflectCAP reaches the Pareto frontier of the trade-off between factuality and coverage, and delivers substantial gains on CapArena-Auto, where generated captions are judged head-to-head against strong reference models. Moreover, ReflectCAP offers a more favorable trade-off between caption quality and compute cost than model scaling or existing multi-agent pipelines, which incur 21–36\% greater overhead. This makes high-quality detailed captioning viable under real-world cost and latency constraints.
\end{abstract}
\section{Introduction}

\emph{Hyper-detailed captions} capture not only salient objects but also their attributes, orientations, spatial relations, background context, and subtle visual states, forming a comprehensive textual representation of an image.
Such captions have become a key ingredient in downstream multimodal systems—improving prompt fidelity for text-to-image and text-to-video generation, and serving as a reasoning aid for compositional and grounded decision-making \cite{betker2023improving,gutflaish2025generating,merchant2025structuredcaptionsimproveprompt,brooks2024video,ju2024miradata,garg-etal-2024-imageinwords}. Large vision-language models (LVLMs) can produce these descriptions fluently \cite{liu2023visual,zhu2023minigpt,dai2023instructblip,liu2024improved}, yet they frequently hallucinate—generating text that is not grounded in the image. This limitation is widely attributed to the tendency of language priors to progressively dominate over visual evidence as generation length increases, leading the model to describe what is statistically probable rather than what is actually depicted \cite{min-etal-2025-mitigating,lee2025toward,lee-etal-2025-vlind,liu2023mitigating}. Hyper-detailed captioning, which inherently requires such extended generation, thus remains a critical bottleneck for reliable deployment.

The straightforward remedy is supervised fine-tuning on human-authored detailed captions \cite{garg-etal-2024-imageinwords,onoe2024doccidescriptionsconnectedcontrasting} to improve LVLMs' intrinsic performance; however, as we demonstrate in Section~\ref{sec:sft}, such captions often exceed the model's perceptual capacity, even increasing hallucinations well beyond the base model.
Moreover, this approach requires not only expensive human annotation but also additional training, further limiting its practicality.
An alternative is inference-time correction, where the model iteratively revises its own output without additional training—an approach that has proven effective for LLMs \cite{kamoi-etal-2024-llms,madaan2023selfrefineiterativerefinementselffeedback}.
However, recent studies show that LVLMs struggle to self-correct during inference without external feedback, as they tend to confirm rather than rectify their own errors\cite{he2025self,zhang2025sc}.
Moreover, iterative revision lengthens the text context, further amplifying language-prior reliance over visual evidence \cite{min-etal-2025-mitigating,lee2025toward}.
Taken together, these limitations suggest that a single LVLM alone is unlikely to fully address the detail–faithfulness tension, and that external guidance from a multi-agent system is needed.

\begin{figure*}[t]
\centering
\includegraphics[width=0.95\textwidth]{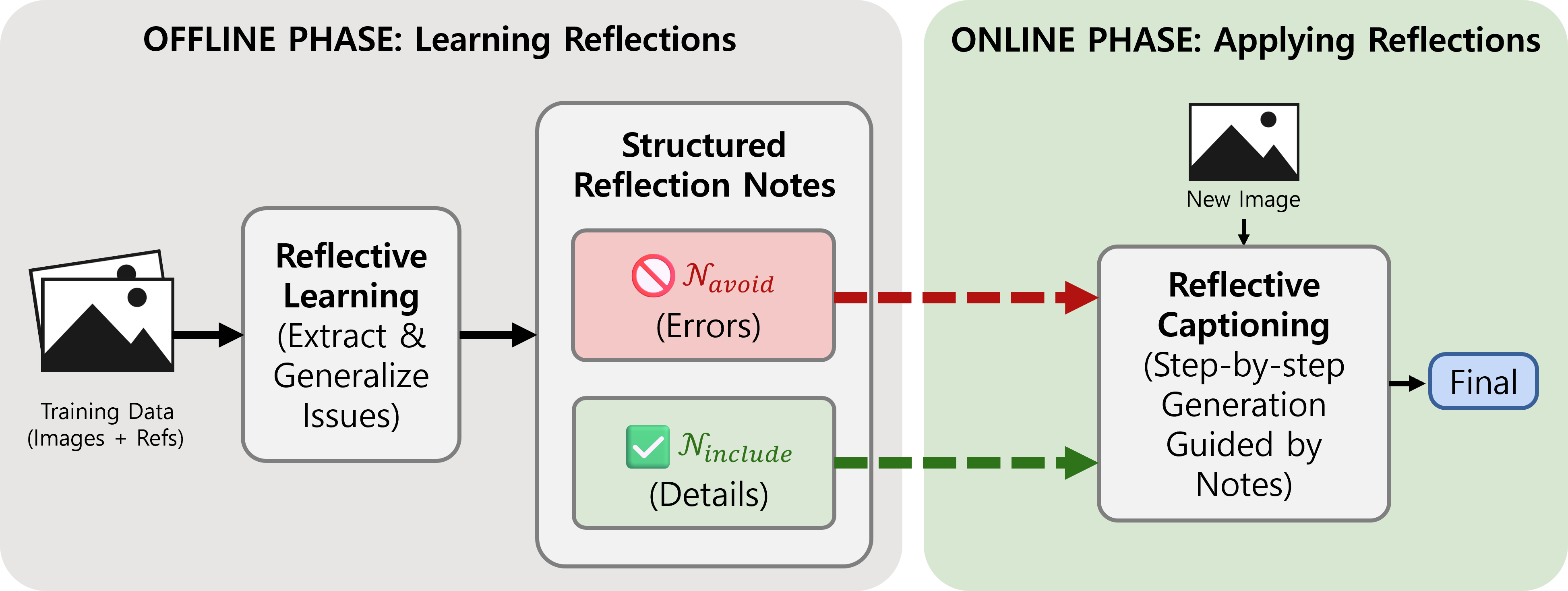}
\caption{Overview of ReflectCAP. In the offline phase, a multi-agent reflective learning pipeline distills a target LVLM’s recurring captioning errors and omissions into Structured Reflection Notes. In the online phase, these notes guide caption generation for new images, producing captions that better balance factuality and coverage.}
\label{fig:intro}
\end{figure*}

To this end, we propose \textbf{ReflectCAP} (Reflective Note-Guided Captioning), a gradient-free framework that distills a target LVLM's recurring errors into an agentic memory called \emph{Structured Reflection Notes} and leverages them for inference-time steering (Figure~\ref{fig:intro}). ReflectCAP operates in two distinct phases.
In the offline phase, a multi-agent pipeline critiques the target model's captions against a small set of human-annotated references to diagnose its systematic error patterns, separately encoding 1) \textbf{hallucination patterns} and 2) \textbf{omission patterns} into generalized guideline notes. In the online phase, each set of notes serves a distinct role in steering generation: one produces a grounded base caption that suppresses recurring hallucinations, and the other produces a detail-focused caption covering typically neglected visual elements. A final merge step combines both with the image as a reference, producing a comprehensive caption that improves factuality and coverage simultaneously.

Empirically, ReflectCAP sets new pareto frontier under the factuality–coverage evaluation proposed by Lee et al.~\cite{lee2025toward}, substantially expanding coverage while preserving precision and achieving the highest F1 across model families. This improvement also holds on CapArena-Auto~\cite{cheng-etal-2025-caparena}, a pairwise benchmark that evaluates captions holistically and is closely aligned with human preference. On 600 images evaluated with CapArena-Auto, ReflectCAP yields substantial win-rate improvements of +32.2 and +21.9 points over the corresponding baselines for the GPT and open-source model families, respectively.
Furthermore, ReflectCAP can elevate smaller models beyond flagship baselines; for instance, GPT-4.1-mini with our method surpasses GPT-5.2 in CapArena-Auto win rate.
Beyond performance comparisons, ReflectCAP consistently offers a more compute-efficient path to improving caption quality than either scaling model size or scaling inference-time computation; for example, InternVL3.5-4B with ReflectCAP achieves a comparable factuality–coverage F1 to InternVL3.5-38B while requiring approximately 8 times lower inference TFLOPs, and outperforms existing multi-agent pipelines while incurring 21–36\% lower compute overhead, suggesting that ReflectCAP can serve as a practical, cost-effective alternative to both model scaling and inference-time computation scaling for detailed captioning.

\section{Related Work}
\label{related_work}

\subsection{Detailed Image Captioning}
Obtaining large-scale detailed image captions is increasingly important, as they serve as training signals for text-to-image and text-to-video generation and as reasoning aids for compositional vision--language tasks \cite{betker2023improving,gutflaish2025generating,brooks2024video,garg-etal-2024-imageinwords}. 
Approaches to scaling detailed captions broadly follow two directions. 
The first relies on human-authored dense captions (e.g., DCI \cite{urbanek2024pictureworth77text}, DOCCI \cite{onoe2024doccidescriptionsconnectedcontrasting}, IIW \cite{garg-etal-2024-imageinwords}), which offer high fidelity and coverage but are prohibitively expensive to scale beyond limited datasets.
The second approach employs LVLMs as automated captioners.
However, in long-form generation, these models often over-rely on language priors, 
which results in hallucinated but unsupported details \cite{min-etal-2025-mitigating,lee-etal-2025-vlind} and the omission of subtle visual attributes  \cite{fu2024blink,rahmanzadehgervi2024vision,marsili2025same}.
Many existing mitigation methods primarily improve factuality (precision) without comparably improving descriptive coverage (recall), leaving the two principal quality axes of detailed captioning in tension\cite{zhou2023analyzing,leng2024mitigating,huang2024opera,favero2024multi,wang2024mitigating,zhu2025ibd}. 
Our work addresses this trade-off by distilling reflection notes that explicitly capture recurring hallucination patterns and missing-detail patterns, enabling separate control of hallucination suppression and detail recovery at inference time.

\subsection{Reflective Memory in Agentic Frameworks}
Recent advancements in LLM-based agents have successfully leveraged reflective memory to refine behavior in long-horizon tasks. By analyzing historical trajectories and inference-time feedback, these text-based models synthesize reusable reasoning strategies and deploy them at appropriate moments to guide subsequent multi-step decision-making\cite{tan2025prospect,zhu2026toward,shinn2023reflexion,wan2025compassenhancingagentlonghorizon,ouyang2025reasoningbankscalingagentselfevolving}.

However, extending this long-horizon reflection paradigm to LVLMs is fundamentally problematic. Unlike pure text generation, LVLMs struggle to reliably extract error patterns over extended reasoning chains\cite{he2025self,zhang2025sc}. As the number of inference steps increases, the visual evidence becomes increasingly diluted, and the models fall prey to severe language prior phenomena—relying more on text-induced hallucinations than on the grounded visual input \cite{li2025the,sun-etal-2025-mitigating-visual,min-etal-2025-mitigating,chung2026v1learningpointvisual}

To address these intrinsic limitations, we shift from online, long-horizon trajectory tracking to an offline, bottom-up distillation process. By systematically aggregating image-specific feedback across diverse samples, we identify and distill the target LVLM's recurring hallucination and omission patterns into a generalized reflection memory. This distilled memory is then utilized to proactively steer the model during inference, bypassing the need for costly step-by-step refinement.

\section{Reflective Note-Guided Captioning Framework}
\label{sec:method}

\begin{figure*}[t]
\centering
\includegraphics[width=0.95\textwidth]{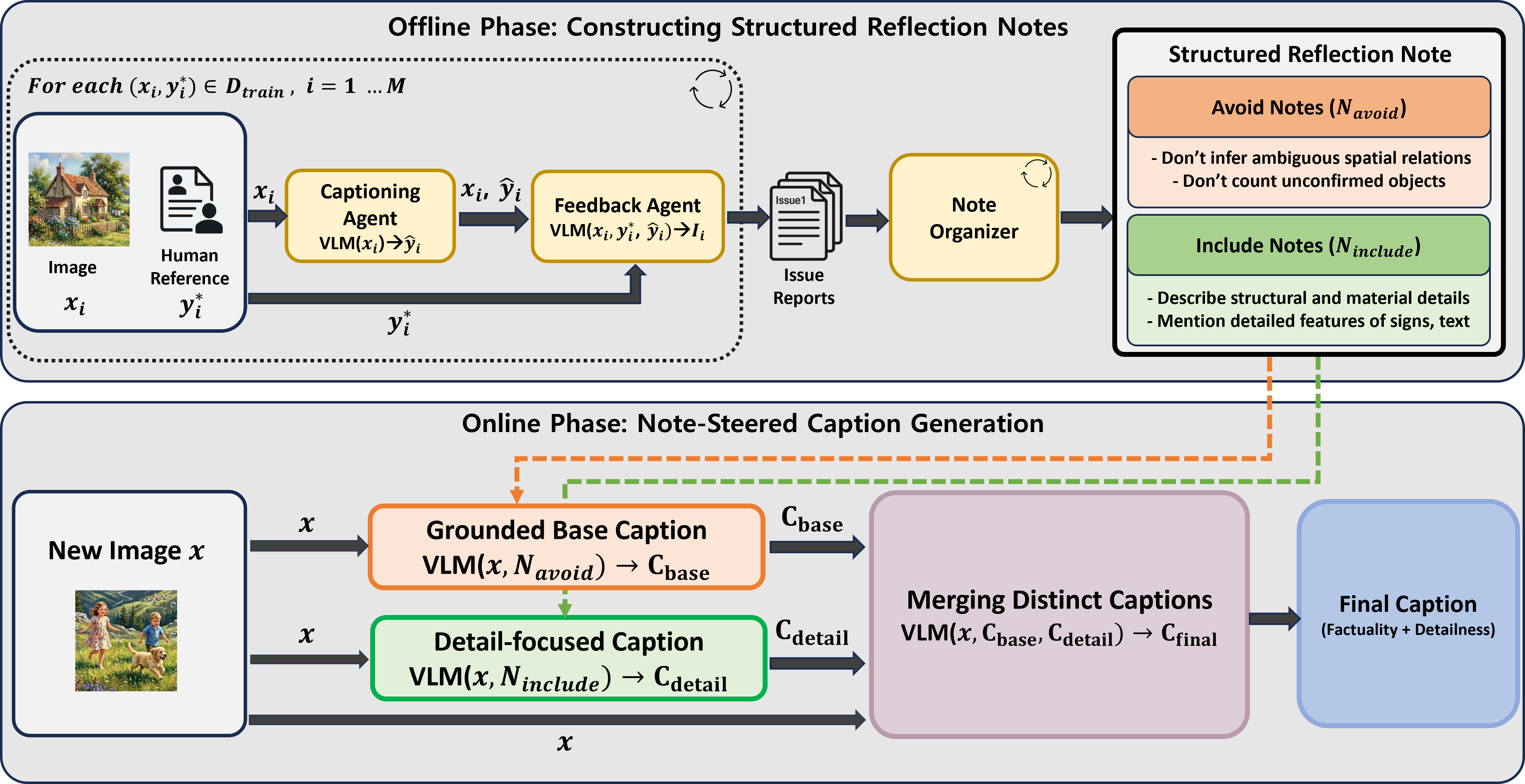}
\caption{
ReflectCAP framework. In the \textbf{offline phase}, a multi-agent pipeline analyzes a small exemplar set to distill recurring errors and omissions of the target LVLM into \emph{Structured Reflection Notes}. In the \textbf{online phase}, these notes guide caption generation: Avoid Notes suppress hallucinations, Include Notes encourage missing details, and a final merge integrates grounded and detail-focused captions into the final output.}
\label{fig:method_pipe}
\end{figure*}

We introduce Reflective Note-Guided Captioning (ReflectCAP), a framework that 1) distills systematic error patterns of a target LVLM into compact, reusable directives, and 2) leverages them as guidance to improve both factuality and detailedness in image captioning.
The key insight is that large vision-language models exhibit predictable failure modes, including recurring hallucinations and consistent blind spots. Once surfaced, these patterns can be counteracted through lightweight prompt-level intervention rather than costly retraining or multi-agent inference at test time. ReflectCAP operationalizes this insight in two stages: an \textbf{offline phase} that analyzes a small exemplar set through a multi-agent pipeline to construct structured reflection notes encoding the model's characteristic errors, and an \textbf{online phase} that injects these notes into the generation context for new images, achieving improved factuality and coverage at negligible additional cost. Figure~\ref{fig:method_pipe} illustrates the overall pipeline.

\subsection{Offline Phase: Constructing Structured Reflection Notes}
\label{sec:online}

The goal of the offline phase is to surface the target LVLM's systematic error patterns and encode them as reusable guidance. Given a small exemplar set $\mathcal{D}_{\text{train}} = \{(x_i, y_i^*)\}_{i=1}^{M}$ of images paired with human-written reference captions, where $x_i$ denotes the $i$-th input image, $y_i^*$ its corresponding reference caption, and $M$ the total number of exemplars, we run a three-agent pipeline that progressively moves from raw errors to generalizable directives.

\noindent\textbf{Captioning Agent.} The pipeline begins by letting the target LVLM caption each image $x_i$ in a zero-shot manner, producing a candidate caption $\hat{y}_i$ with no additional guidance. This is intentional: the resulting captions faithfully reflect the model's default behavior---including its characteristic hallucinations and omissions---providing an unbiased basis for the diagnosis that follows.

\noindent\textbf{Feedback Agent.} Each candidate caption is then critiqued against two sources of evidence: the image itself and the human reference. The Feedback Agent cross-references $\hat{y}_i$ against both $x_i$ and $y_i^*$, producing a structured issue report $\mathcal{I}_i$ for each example. Reports are organized into two categories:
\begin{itemize}
    \item \textbf{Hallucinations}: details in $\hat{y}_i$ that are factually incorrect or not visible in $x_i$.
    \item \textbf{Missing Details}: important details present in $y_i^*$ that are absent from $\hat{y}_i$.
\end{itemize}
For example, a hallucination might be \textit{``the caption states two people are sitting, but the image shows three,''} while a missing detail might be \textit{``the caption does not mention the wooden railing visible in the foreground.''} The Feedback Agent has access to the image during critique, ensuring that its judgments are visually grounded. However, at this stage, every diagnosis is tied to a particular image and caption, making it difficult to apply directly at inference time.

\noindent\textbf{Note Organizer.} Instance-specific diagnoses help explain individual failures, but reusable guidance requires identifying \emph{patterns}—errors that recur across images. The Note Organizer performs this consolidation.
Because diagnoses collected across images can easily exceed the LVLM context window, the organizer processes them incrementally, consuming batches and updating a running set of notes after each step. During each update, it merges semantically similar issues and abstracts them into broadly applicable rules. By imposing an upper bound of
$K$ items, the note set retains only patterns corresponding to frequently recurring mistakes or commonly omitted details, thereby pruning redundant or overly narrow entries.
The result is a compact, prioritized set of notes that we call \textbf{Structured Reflection Notes}. It consists of two complementary components:

\begin{itemize}
    \item \textbf{Avoid Notes} $\mathcal{N}_{\text{avoid}}$: directives that suppress recurrent hallucination patterns (e.g., \textit{``Do not infer object colors when they are ambiguous''}).
    \item \textbf{Include Notes} $\mathcal{N}_{\text{include}}$: directives that enforce frequently omitted details (e.g., \textit{``Describe visible architectural details such as structural supports and railings''}).
\end{itemize}
This progression from instance-level diagnosis to cross-instance generalization is what allows the notes to capture the model's systematic tendencies rather than one-off mistakes. In practice, $M{=}30$ exemplar images and $K{=}5$ items per category are sufficient, as shown in our ablation study (\S\ref{sec:ablation}). Algorithm~\ref{alg:offline} summarizes the full offline procedure.

\begin{algorithm}[t]
\caption{Offline: Constructing Structured Reflection Notes}
\label{alg:offline}
\begin{algorithmic}[1]
\REQUIRE Training set $\mathcal{D}_{\text{train}} = \{(x_i, y_i^*)\}_{i=1}^{M}$, max items $K$, batch size $B$
\STATE $\mathcal{N} \leftarrow \emptyset$
\FOR{$i = 1$ to $M$}
    \STATE $\hat{y}_i \leftarrow \text{CaptioningAgent}(x_i)$ \hfill $\triangleright$ Zero-shot captioning
    \STATE $\mathcal{I}_i \leftarrow \text{FeedbackAgent}(x_i, \hat{y}_i, y_i^*)$ \hfill $\triangleright$ Instance-specific critique
\ENDFOR
\FOR{each batch $\mathcal{B} \subset \{\mathcal{I}_1, \ldots, \mathcal{I}_M\}$ of size $B$}
    \STATE $\mathcal{N} \leftarrow \text{NoteOrganizer}(\mathcal{B}, \mathcal{N}, K)$ \hfill $\triangleright$ Cross-instance generalization
\ENDFOR
\RETURN $\mathcal{N} = (\mathcal{N}_{\text{avoid}}, \mathcal{N}_{\text{include}})$
\end{algorithmic}
\end{algorithm}

\begin{algorithm}[t]
\caption{Online: Note-Steered Caption Generation}
\label{alg:online}
\begin{algorithmic}[1]
\REQUIRE Test image $x$, Structured Reflection Notes $\mathcal{N} = (\mathcal{N}_{\text{avoid}}, \mathcal{N}_{\text{include}})$
\STATE \textbf{Step 1: Grounded Base Caption}
\STATE $c_{\text{base}} \leftarrow \text{VLM}(x,\; \mathcal{N}_{\text{avoid}})$ \hfill $\triangleright$ Suppress known hallucination patterns
\STATE \textbf{Step 2: Detail-Focused Caption}
\STATE $c_{\text{detail}} \leftarrow \text{VLM}(x,\; \mathcal{N}_{\text{include}})$ \hfill $\triangleright$ Attend to typically neglected details
\STATE \textbf{Step 3: Merging Distinct Captions}
\STATE $c_{\text{final}} \leftarrow \text{VLM}(x,\; c_{\text{base}},\; c_{\text{detail}})$ \hfill $\triangleright$ Merge with image as reference
\RETURN $c_{\text{final}}$
\end{algorithmic}
\end{algorithm}

\subsection{Online Phase: Note-Steered Caption Generation}
\label{sec:online}

Once constructed, the Structured Reflection Notes replace the multi-agent pipeline entirely. Given a new image $x$ and the pre-computed notes $\mathcal{N} = (\mathcal{N}_{\text{avoid}}, \mathcal{N}_{\text{include}})$, the online phase generates a caption through at most three LVLM calls.

\noindent\textbf{Step 1: Grounded Base Caption.} $\mathcal{N}_{\text{avoid}}$ is injected into the captioning prompt, directing the LVLM to suppress its known hallucination patterns during generation. This produces a grounded base caption $c_{\text{base}}$ that is more factually reliable than a zero-shot caption while preserving the model's natural descriptive ability. Since this step requires exactly one forward pass, almost identical in cost to zero-shot inference, it can serve as a standalone variant for captioning pipelines where factuality is the primary concern.

\noindent\textbf{Step 2: Detail-Focused Caption.} A second call uses $\mathcal{N}_{\text{include}}$ to direct the LVLM's attention toward the specific types of details it typically misses, such as material textures, background elements, and spatial arrangements, producing a detail-focused caption $c_{\text{detail}}$. This caption captures the descriptive details that $c_{\text{base}}$ trades off in favor of factual grounding.

\noindent\textbf{Step 3: Merging Distinct Captions.} The final stage merges $c_{\text{base}}$ and $c_{\text{detail}}$ into a unified caption $c_{\text{final}}$, using the image as a grounding reference. To prevent the integration of spurious details, the merge strategy is conservative: $c_{\text{base}}$ is treated as the primary source of truth, while $c_{\text{detail}}$ serves as a supplementary source. In cases of conflict, the model prioritizes $c_{\text{base}}$, as it is generated under hallucination-suppressing guidance designed for factual grounding. We refer to this full three-step pipeline as \textbf{ReflectCAP}. The complete procedure is summarized in Algorithm~\ref{alg:online}.

\section{Experiments}
\label{sec:experiments}

We evaluate ReflectCAP along three dimensions. (1)~\textbf{Fine-grained evaluation}: fine-grained factuality and coverage, which examines the trade-off between these two axes, (2)~\textbf{Holistic evaluation}: holistic caption quality, which tests whether fine-grained gains translate into perceived overall quality, and (3)~\textbf{Cost-efficiency}: computational cost analysis which examines whether the method is practical enough for downstream deployment.

\subsection{Experimental Settings}
\label{sec:settings}

\noindent\textbf{Evaluation Metrics.} We adopt two complementary evaluation suites, both well-aligned with human judgments. For fine-grained evaluation, we evaluate on IIW-400 dataset~\cite{garg-etal-2024-imageinwords} using the factuality and coverage metrics proposed by Lee et al.~\cite{lee2025toward}. \emph{Factuality} (Precision) decomposes each caption into atomic propositions and verifies each against the image and ground-truth; \emph{Coverage} (Recall) is measured via curated VQA items associated with each IIW-400 image, answered using only the generated caption. We report \emph{F1 score} as the harmonic mean of factuality and coverage. For holistic evaluation, we adopt CapArena-Auto~\cite{cheng-etal-2025-caparena}, a pairwise benchmark scored by average win-rate margin against three reference models.

\noindent\textbf{Baseline  LVLMs.} We evaluate across eight LVLMs spanning closed-source and open-source families at a range of scales: two proprietary models (GPT-4.1-mini, GPT-4.1-nano) and six open-weight instruction-tuned models, comprising three smaller models (InternVL3.5-4B-Instruct, Qwen2.5-VL-7B-Instruct, Qwen3-VL-8B-Instruct) and three larger models (InternVL3.5-38B-Instruct, Qwen2.5-VL-32B-Instruct, Qwen3-VL-32B-Instruct).

\noindent\textbf{Baseline Methods.} For each model, we compare ReflectCAP against four captioning strategies. \emph{Zero-shot} uses a minimal prompt (``Describe this image in detail''). \emph{Few-shot} prepends three randomly sampled human-annotated caption exemplars. \emph{Self-Correction} first generates a zero-shot caption, then revises it after re-examining the image. \emph{CapMAS}~\cite{lee2025toward}, a multi-agent baseline, decomposes a caption into atomic propositions via specialized agents, verifies each against the image, and rewrites the caption by removing unverified content. For fair comparison, ReflectCAP uses the target model itself for both the offline phase (note construction) and the online phase (caption generation), ensuring that no external model contributes to the final output. In the offline phase, we use images and reference captions from IIW-Eval that are not included in IIW-400 to construct the notes.

\begin{table*}[t]
\centering
\caption{\textbf{Factuality and Coverage on IIW-400.} Precision measures the ratio of verified-true propositions. Recall measures the ratio of correctly answered VQA questions. F1 is the harmonic mean. Best per model in \textbf{bold}. $\Delta$ denotes F1 change from Zero-shot.}
\label{tab:main}
\setlength{\tabcolsep}{2pt}
\renewcommand{\arraystretch}{1}
\footnotesize
\begin{adjustbox}{max width=0.95\textwidth}
\begin{tabular}{@{}cl cccc cl cccc@{}}
\toprule
\textbf{Model} & \textbf{Method} & \textbf{P} & \textbf{R} & \textbf{F1} & $\boldsymbol{\Delta}$
& \textbf{Model} & \textbf{Method} & \textbf{P} & \textbf{R} & \textbf{F1} & $\boldsymbol{\Delta}$ \\
\midrule
\multirow{5}{*}{\rotatebox{90}{\scriptsize GPT-4.1-mini}}
 & Zero-shot   & 83.6 & 68.1 & 75.1 & ---
 & \multirow{5}{*}{\rotatebox{90}{\scriptsize Qwen2.5-7B}}
 & Zero-shot   & 68.3 & 57.9 & 62.7 & --- \\
 & Few-shot    & 81.9 & 71.5 & 76.3 & +1.2
 && Few-shot    & 49.9 & 56.6 & 53.1 & $-$9.6 \\
 & Self-Corr.  & 82.6 & 69.3 & 75.4 & +0.3
 && Self-Corr.  & 63.8 & 57.3 & 60.4 & $-$2.3 \\
 & CapMAS      & \textbf{84.2} & 67.7 & 75.1 & 0.0
 && CapMAS      & \textbf{72.0} & 57.4 & 63.9 & +1.2 \\
 & \hl{\textbf{ReflectCAP}} & \hl{83.8} & \hl{\textbf{72.0}} & \hl{\textbf{77.5}} & \hl{\textbf{+2.4}}
 && \hl{\textbf{ReflectCAP}} & \hl{68.8} & \hl{\textbf{62.3}} & \hl{\textbf{65.4}} & \hl{\textbf{+2.7}} \\
\midrule
\multirow{5}{*}{\rotatebox{90}{\scriptsize GPT-4.1-nano}}
 & Zero-shot   & 78.9 & 62.6 & 69.8 & ---
 & \multirow{5}{*}{\rotatebox{90}{\scriptsize Qwen2.5-32B}}
 & Zero-shot   & 71.9 & 64.0 & 67.7 & --- \\
 & Few-shot    & 77.1 & 67.1 & 71.8 & +2.0
 && Few-shot    & 64.9 & 64.5 & 64.7 & $-$3.0 \\
 & Self-Corr.  & 79.2 & 62.6 & 70.0 & +0.2
 && Self-Corr.  & 70.2 & 66.2 & \textbf{68.1} & \textbf{+0.4} \\
 & CapMAS      & \textbf{83.1} & 57.2 & 67.8 & $-$2.0
 && CapMAS      & \textbf{73.5} & 63.1 & 67.9 & +0.2 \\
 & \hl{\textbf{ReflectCAP}} & \hl{78.0} & \hl{\textbf{67.2}} & \hl{\textbf{72.2}} & \hl{\textbf{+2.4}}
 && \hl{\textbf{ReflectCAP}} & \hl{69.9} & \hl{\textbf{66.4}} & \hl{\textbf{68.1}} & \hl{\textbf{+0.4}} \\
\midrule
\multirow{5}{*}{\rotatebox{90}{\scriptsize InternVL-4B}}
 & Zero-shot   & 64.1 & 54.9 & 59.1 & ---
 & \multirow{5}{*}{\rotatebox{90}{\scriptsize Qwen3-8B}}
 & Zero-shot   & 76.3 & 69.2 & 72.6 & --- \\
 & Few-shot    & 49.4 & 52.8 & 51.1 & $-$8.0
 && Few-shot    & 71.3 & \textbf{72.2} & 71.8 & $-$0.8 \\
 & Self-Corr.  & 62.1 & 54.1 & 57.8 & $-$1.3
 && Self-Corr.  & 76.3 & 70.0 & 73.0 & +0.4 \\
 & CapMAS      & \textbf{72.2} & 53.3 & 61.3 & +2.2
 && CapMAS      & \textbf{79.3} & 68.8 & 73.7 & +1.1 \\
 & \hl{\textbf{ReflectCAP}} & \hl{64.8} & \hl{\textbf{61.3}} & \hl{\textbf{63.0}} & \hl{\textbf{+3.9}}
 && \hl{\textbf{ReflectCAP}} & \hl{77.0} & \hl{71.2} & \hl{\textbf{74.0}} & \hl{\textbf{+1.4}} \\
\midrule
\multirow{5}{*}{\rotatebox{90}{\scriptsize InternVL-38B}}
 & Zero-shot   & 72.5 & 57.8 & 64.3 & ---
 & \multirow{5}{*}{\rotatebox{90}{\scriptsize Qwen3-32B}}
 & Zero-shot   & 79.2 & 72.5 & 75.7 & --- \\
 & Few-shot    & 63.2 & 62.1 & 62.6 & $-$1.7
 && Few-shot    & 73.4 & \textbf{74.4} & 73.9 & $-$1.8 \\
 & Self-Corr.  & 73.7 & 58.2 & 65.0 & +0.7
 && Self-Corr.  & 78.6 & 73.6 & 76.0 & +0.3 \\
 & CapMAS      & \textbf{78.6} & 57.5 & 66.4 & +2.1
 && CapMAS      & \textbf{81.0} & 72.0 & 76.2 & +0.5 \\
 & \hl{\textbf{ReflectCAP}} & \hl{73.7} & \hl{\textbf{64.2}} & \hl{\textbf{68.6}} & \hl{\textbf{+4.3}}
 && \hl{\textbf{ReflectCAP}} & \hl{79.2} & \hl{73.8} & \hl{\textbf{76.4}} & \hl{\textbf{+0.7}} \\
\bottomrule
\end{tabular}
\end{adjustbox}
\end{table*}

\subsection{Fine-Grained Evaluation}
\label{sec:fine_grained}

Table~\ref{tab:main} presents our main results. Existing methods tend to improve one axis at the cost of the other: few-shot prompting increases coverage by imitating human-authored demonstrations but pushes the model beyond its perceptual boundary, causing factuality to drop. Self-correction yields only marginal gains regardless of model scale, indicating that models struggle to identify and fix errors in their own generated captions through revision alone. CapMAS that improves factuality by removing unreliable content from existing captions inevitably sacrifices coverage in the process. In contrast, ReflectCAP substantially boosts coverage while incurring minimal loss in factuality, achieving the highest $F_1$ across all eight models. This demonstrates that the Structured Reflection Notes effectively balance the inherent trade-off: coverage guidance encourages the model to describe more, which inevitably risks lowering factuality, while factuality guidance separately constrains this degradation.
By controlling each objective through its own dedicated guidance, ReflectCAP achieves the highest $F_1$ across all models, advancing the Pareto frontier between the two objectives.

\begin{table}[t]
\centering
\caption{\textbf{CapArena-Auto scores.} Score denotes the average win-rate margin (higher is better; range $[-100,100]$). For CapMAS and ReflectCAP, we report \emph{score} with the change relative to zero-shot in parentheses. Rows marked with $\dagger$ use zero-shot reference values taken from the provided CapArena caption.}
\label{tab:caparena}
\adjustbox{max width=0.9\textwidth}{%
\footnotesize
\begin{tabular}{@{}l c c c@{}}
\toprule
\textbf{Model} & \textbf{Zero-shot} & \textbf{CapMAS} & \textbf{ReflectCAP (Ours)} \\
\midrule
\multicolumn{4}{@{}l}{\textit{Leaderboard anchors (zero-shot only)}} \\
GPT-5.2 & 70.0 & --- & --- \\
Gemini-1.5-Pro$^\dagger$    & 62.3 & --- & --- \\
GPT-4o-0806$^\dagger$       & 44.3 & --- & --- \\
Qwen2.5VL-72B$^\dagger$     & 39.7 & --- & --- \\
Claude-3.5-Sonnet$^\dagger$ & 29.7 & --- & --- \\
\midrule
GPT-4.1-mini
    & 57.7
    & 53.7 ($-4.0$)
    & \textbf{90.0} ($+32.3$) \\
GPT-4.1-nano
    & 21.2
    & $-14.7$ ($-35.9$)
    & \textbf{51.3} ($+30.1$) \\
InternVL3.5-4B
    & $-54.3$
    & $-46.7$ ($+7.6$)
    & \bm{$-18.7$} ($+35.6$) \\
InternVL3.5-38B
    & $-24.0$
    & $-15.0$ ($+9.0$)
    & \textbf{12.0} ($+36.0$) \\
Qwen2.5-VL-7B
    & $-33.7$
    & $-28.0$ ($+5.7$)
    & \bm{$-7.3$} ($+26.4$) \\
Qwen2.5-VL-32B
    & 5.7
    & 8.0 ($+2.3$)
    & \textbf{25.3} ($+19.6$) \\
Qwen3-VL-8B
    & 76.0
    & 74.0 ($-2.0$)
    & \textbf{85.3} ($+9.3$) \\
Qwen3-VL-32B
    & 87.3
    & 87.0 ($-0.3$)
    & \textbf{91.7} ($+4.4$) \\
\bottomrule
\end{tabular}
}%
\end{table}

\subsection{Holistic Evaluation}
\label{sec:holistic}

Fine-grained metrics measure factuality and coverage in isolation, but it is also necessary to evaluate overall caption quality.
CapArena-Auto (Table~\ref{tab:caparena}) tests this by pitting each method's captions against fixed three reference models\footnote{GPT-4o-0806, CogVLM2-llama3-chat-19B, and MiniCPM-V2.6-8B.} in head-to-head comparisons judged by GPT-4.1-mini.

ReflectCAP improves the average win-rate margin by +33.2 points for the GPT family and +21.9 points for open-source models over their zero-shot baselines, confirming that the fine-grained gains in \S\ref{sec:fine_grained} translate directly into holistic caption quality. 
Notably, GPT-4.1-mini with ReflectCAP (90.0) surpasses even GPT-5.2 (70.0), suggesting that structured reflection notes can compensate for inherent model capacity differences in overall caption quality.
Additionally, ReflectCAP yields consistent positive gains over the zero-shot baseline across all models, whereas CapMAS tends to degrade performance when applied to models that already exhibit strong zero-shot capabilities. 

\begin{figure*}[t]
\centering
\includegraphics[width=0.95\textwidth]{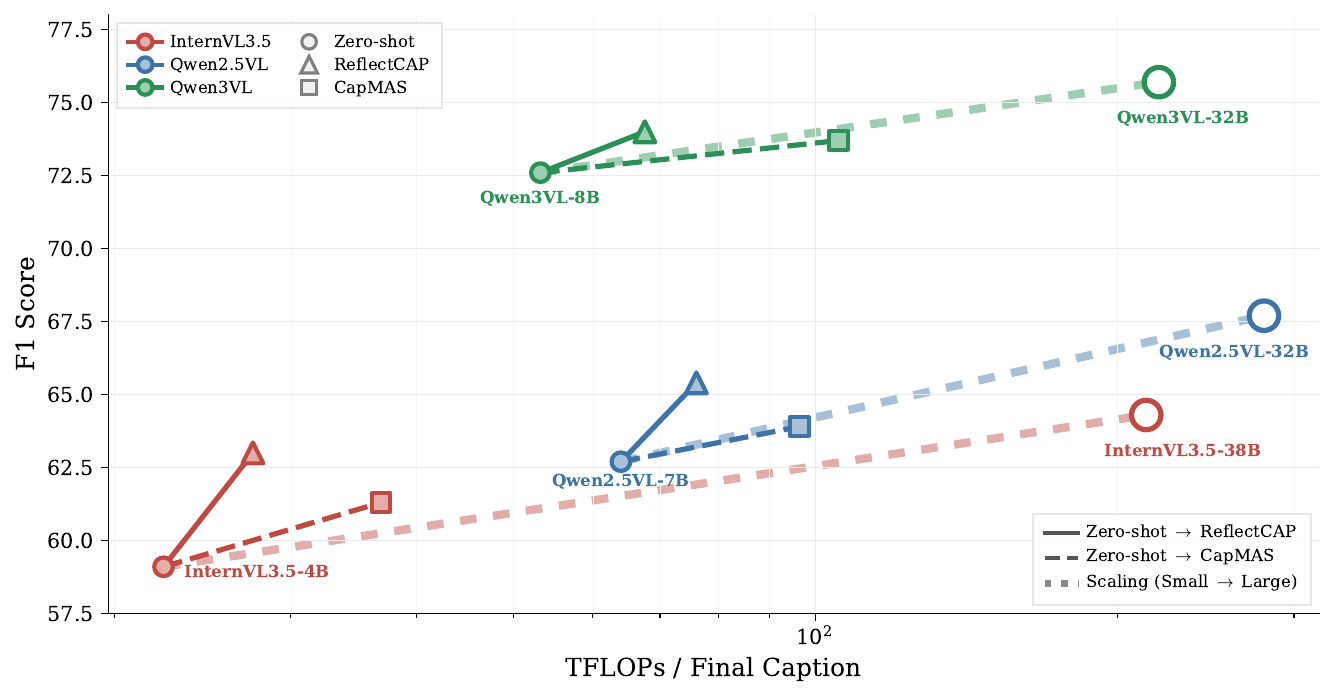}
\caption{\textbf{Solid} and \textbf{dash-dotted lines} denote improvements from zero-shot to ReflectCAP and CapMAS, respectively. ReflectCAP achieves higher F1 scores while requiring 21–36\% less compute than CapMAS. \textbf{Light dashed lines} denote performance gains from model parameter scaling. Compared to simply increasing model size, ReflectCAP achieves comparable quality at up to $8\times$ lower compute cost, enabling high-quality, detailed captioning more practical under real-world cost and latency constraints.}
\label{fig:cost}
\end{figure*}

\subsection{Cost-Efficiency}
\label{sec:cost}

Beyond caption quality, practical deployment requires efficient inference. To examine this, we measure cost-efficiency across methods and open-source models in terms of the total TFLOPs required to produce a final caption at inference time.\footnote{We approximate inference cost as $C \approx 2NT$, where $N$ is the number of non-embedding parameters and $T$ is the total token count. For methods with multiple calls per image, image tokens are counted only once via KV caching.}

Figure~\ref{fig:cost} plots $F_1$ against inference TFLOPs per final caption for all open-source models. Two trends emerge.
First, ReflectCAP improves caption quality more compute-efficiently than simply scaling model size. For example, InternVL3.5-4B with ReflectCAP achieves an $F_1$ of 63.0, approaching InternVL3.5-38B zero-shot (64.3) while requiring roughly 7.8× fewer TFLOPs (27.5 vs. 213.7). Similarly, Qwen3-VL-8B with ReflectCAP maintains a comparable performance gap relative to Qwen3-VL-32B zero-shot, while generating captions at approximately 3.3× lower computational cost.
Second, ReflectCAP is also more compute-efficient than existing inference-time baselines within the same model. Across three models— InternVL3.5-4B, Qwen2.5-VL-7B, and Qwen3-VL-8B—ReflectCAP reduces inference cost by 21\%–36\% TFLOPs compared to CapMAS, while consistently achieving higher $F_1$ scores. This inefficiency stems from CapMAS applying its multi-agent pipeline directly at inference time, and its focus on factuality alone limits overall caption quality.
These results highlight ReflectCAP as a practical alternative to both model scaling and compute-heavy inference-time pipelines.

\section{Analysis}
\label{sec:analysis}

\subsection{Supervised Fine-Tuning on Detailed Image Captioning}
\label{sec:sft}

\begin{table}[t]
\centering
\caption{Impact of SFT on detailed captioning factuality.}
\label{tab:sft}
\setlength{\tabcolsep}{4pt}
\renewcommand{\arraystretch}{0.95}
\footnotesize
\begin{tabular}{@{}cl ccc@{}}
\toprule
\textbf{Model} & \textbf{Method} & \textbf{Fact.} & \textbf{Cov.} & \textbf{F1} \\
\midrule
\multirow{3}{*}{{\scriptsize InternVL3.5-4B}}
 & Zero-shot                  & 64.1 & 54.9 & 59.1 \\
 & SFT w/ Human-authored      & 58.2 & 56.5 & 57.4 \\
 & SFT w/ ReflectCAP-generated & \textbf{66.5} & \textbf{61.6} & \textbf{63.9} \\
\midrule
\multirow{3}{*}{{\scriptsize Qwen2.5-VL-7B}}
 & Zero-shot                  & 68.3 & 57.9 & 62.7 \\
 & SFT w/ Human-authored      & 57.1 & 58.0 & 57.6 \\
 & SFT w/ ReflectCAP-generated & \textbf{69.9} & \textbf{63.5} & \textbf{66.5} \\
\bottomrule
\end{tabular}
\end{table}

The most intuitive approach to improving detailed captioning is supervised fine-tuning (SFT) on human-authored detailed captions. To examine this, we fine-tune InternVL3.5-4B and Qwen2.5-VL-7B on 9,647 DOCCI human-annotated captions using LoRA. As shown in Table~\ref{tab:sft}, SFT substantially degrades factuality compared to the zero-shot baseline for both models (evaluation follows the same protocol as Section~\ref{sec:experiments}), lending further support to recent findings that training on annotations exceeding the model's visual capabilities amplifies hallucinations~\cite{yanuka-etal-2025-bridging, yue-etal-2024-less}. An interesting finding is that replacing human captions with ReflectCAP-generated captions on the same DOCCI images for training maintains factuality while improving coverage. This suggests that ReflectCAP can serve as a scalable pipeline for constructing training data that improves recall while staying within the model's visual boundary, without manual annotation effort.

\subsection{Generalization with Frozen Reflection Notes}
\label{sec:frozen_note_generalization}

We further examine whether the reflection notes learned from the original benchmark transfer to external detailed-captioning benchmarks.
Specifically, we evaluate ReflectCAP on CaptionQA~\cite{yang2026captionqacaptionusefulimage} and CAPability~\cite{liu2026capability} using the same frozen IIW-derived notes, without re-mining notes for each target benchmark.
CaptionQA measures downstream caption utility across four domains, while CAPability evaluates caption correctness and thoroughness across multiple captioning dimensions.
For evaluation, we sample 50 images per domain from CaptionQA (4 domains, 200 images total) and 100 images per dimension from CAPability (9 dimensions, 900 images total) using seed 42.

Table~\ref{tab:captionqa} shows that ReflectCAP consistently improves over zero-shot and CapMAS on CaptionQA.
On GPT-4.1-mini, ReflectCAP improves the overall CaptionQA accuracy from 74.5 to 81.8, and on Qwen3-VL-8B, it improves the score from 74.9 to 78.0.
ReflectCAP also improves the average CAPability score on both backbones.
These results suggest that the reflection notes capture reusable model-level error tendencies rather than benchmark-specific artifacts.

\begin{table}[!tbp]
\centering
\scriptsize
\caption{\textbf{Non-homologous generalization.} CaptionQA accuracy (\%; Nat./Doc./E-com/Emb.=natural/document/e-commerce/embodied); CAPability reports P=precision, H=hit, Avg=(P+H)/2 over 9 dims.}
\label{tab:captionqa}
\resizebox{0.99\columnwidth}{!}{%
\begin{tabular}{llccccc|ccc}
\toprule
 & & \multicolumn{5}{c|}{\textbf{CaptionQA}} & \multicolumn{3}{c}{\textbf{CAPability}} \\
\cmidrule(lr){3-7}\cmidrule(lr){8-10}
Backbone & Method & Nat. & Doc. & E-com & Emb. & Ovr. & P & H & Avg \\
\midrule
\multirow{3}{*}{GPT-4.1-mini}
 & ZS     & 70.8 & 81.3 & 83.4 & 64.1 & 74.5 (-) & \textbf{85.2} & 51.9 & 68.6 (-) \\
 & CapMAS & 70.8 & 78.7 & 82.6 & 63.6 & 73.6 (-0.9) & 83.9 & 49.9 & 66.9 (-1.7) \\
 & \textbf{Ours} & \textbf{81.9} & \textbf{84.7} & \textbf{88.4} & \textbf{72.1} & \textbf{81.8} \textbf{(+7.3)} & 83.8 & \textbf{58.2} & \textbf{71.0} \textbf{(+2.4)} \\
\midrule
\multirow{3}{*}{Qwen3-VL-8B}
 & ZS     & 76.1 & 68.7 & 85.0 & 68.3 & 74.9 (-) & 80.9 & 58.1 & 69.5 (-) \\
 & CapMAS & 75.2 & 69.3 & 83.9 & 66.1 & 74.0 (-0.9) & 82.7 & 57.7 & 70.2 (+0.7) \\
 & \textbf{Ours} & \textbf{80.6} & \textbf{74.7} & \textbf{86.1} & \textbf{68.9} & \textbf{78.0} \textbf{(+3.1)} & \textbf{84.2} & \textbf{59.7} & \textbf{71.9} \textbf{(+2.4)} \\
\bottomrule
\end{tabular}}
\end{table}

\subsection{Ablation Study}
\label{sec:ablation}

\begin{figure*}[t]
\centering
\includegraphics[width=0.9\textwidth]{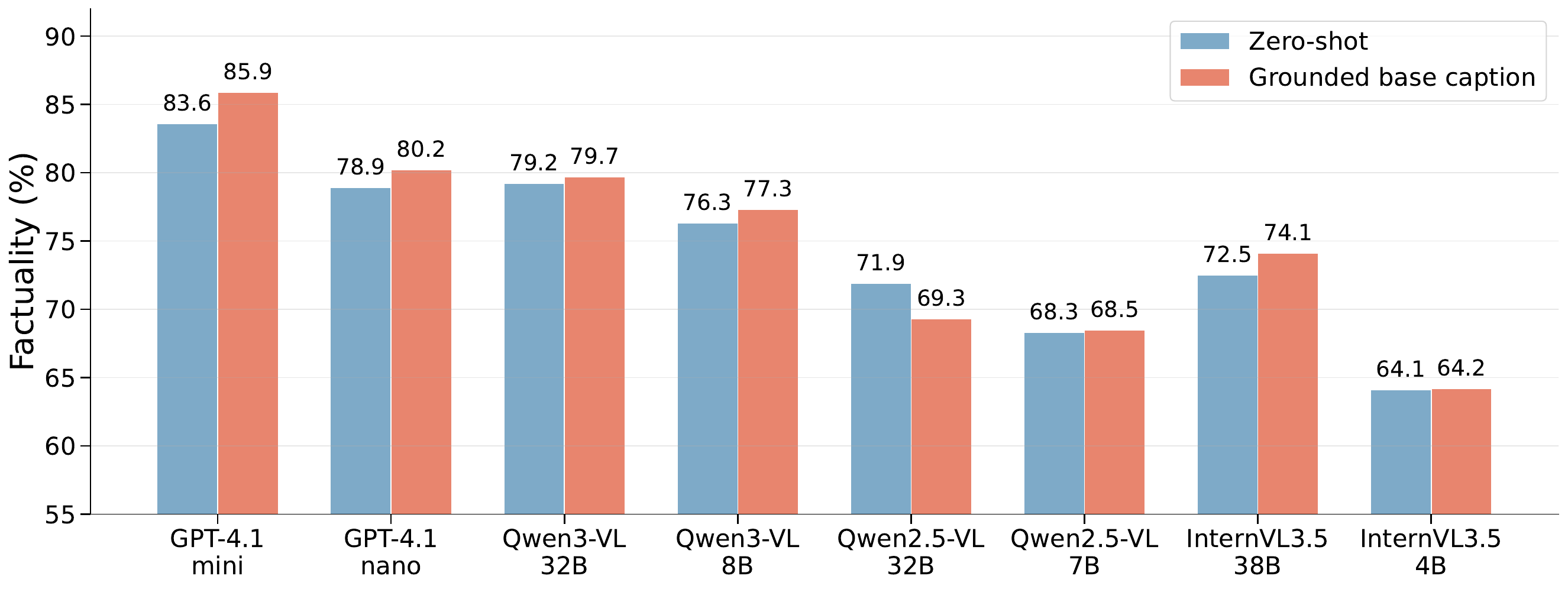}
\caption{Factuality comparison between Zero-shot and Grounded Base Caption across all models. Models with stronger instruction-following capabilities show larger gains. }
\label{fig:factuality_comparison}
\end{figure*}

\noindent\textbf{Effect of Grounded Base Caption.}
As shown in Figure~\ref{fig:factuality_comparison}, the Grounded Base Caption, which applies only the hallucination-suppression notes, improves factuality over the zero-shot baseline in nearly all models. However, the magnitude of this improvement varies with the model's instruction-following capability. Models with stronger instruction-following abilities, such as GPT-4.1-mini and GPT-4.1-nano, generate well-grounded base captions where hallucination patterns are effectively suppressed, whereas the Qwen2.5-VL family shows marginal improvement or even degradation compared to the zero-shot baseline. This suggests that as instruction-following capabilities of LVLMs continue to advance, ReflectCAP can achieve even greater improvements without any modification to the framework.

\noindent\textbf{Separate vs.\ Combined Injection.}
Table~\ref{tab:separate} compares two strategies on 100 images sampled from IIW-400: 
applying Avoid and Include notes in separate generation passes versus injecting both into a single prompt.
Across all four models, the separated approach consistently outperforms the combined variant. This effect is particularly pronounced in InternVL3.5-4B, where the combined injection causes a severe F1 drop (64.4 $\to$ 57.3), indicating that overloading the prompt with too many directives can be detrimental, especially for models with limited instruction-following capability. These results confirm that hallucination suppression and detail recovery are better handled as separate objectives.

\begin{table}[t]
\centering
\caption{Separate vs.\ Combined note injection. Separate-Merge applies Avoid and Include notes in separate passes with merging; Combined injects both into a single prompt.}
\label{tab:separate}
\setlength{\tabcolsep}{3pt}
\scriptsize
\begin{tabular}{llccc}
\toprule
Model & Method & Fact. & Cov. & F1 \\
\midrule
\multirow{2}{*}{GPT-4.1-mini}
  & Separate-Merge & 84.8 & 72.8 & \textbf{78.3} \\
  & Combined & 84.8 & 70.7 & 77.1 \\
\midrule
\multirow{2}{*}{GPT-4.1-nano}
  & Separate-Merge & 78.1 & 68.6 & \textbf{73.1} \\
  & Combined & 78.4 & 67.0 & 72.2 \\
\midrule
\multirow{2}{*}{Qwen3-VL-8B}
  & Separate-Merge & 77.6 & 72.4 & \textbf{74.9} \\
  & Combined & 74.4 & 71.2 & 72.7 \\
\midrule
\multirow{2}{*}{InternVL3.5-4B}
      & Separate-Merge & 66.6 & 62.4 & \textbf{64.4} \\
  & Combined & 55.8 & 58.8 & 57.3 \\
\bottomrule
\end{tabular}
\end{table}

\begin{figure*}[t]
\centering
\includegraphics[width=0.95\textwidth]{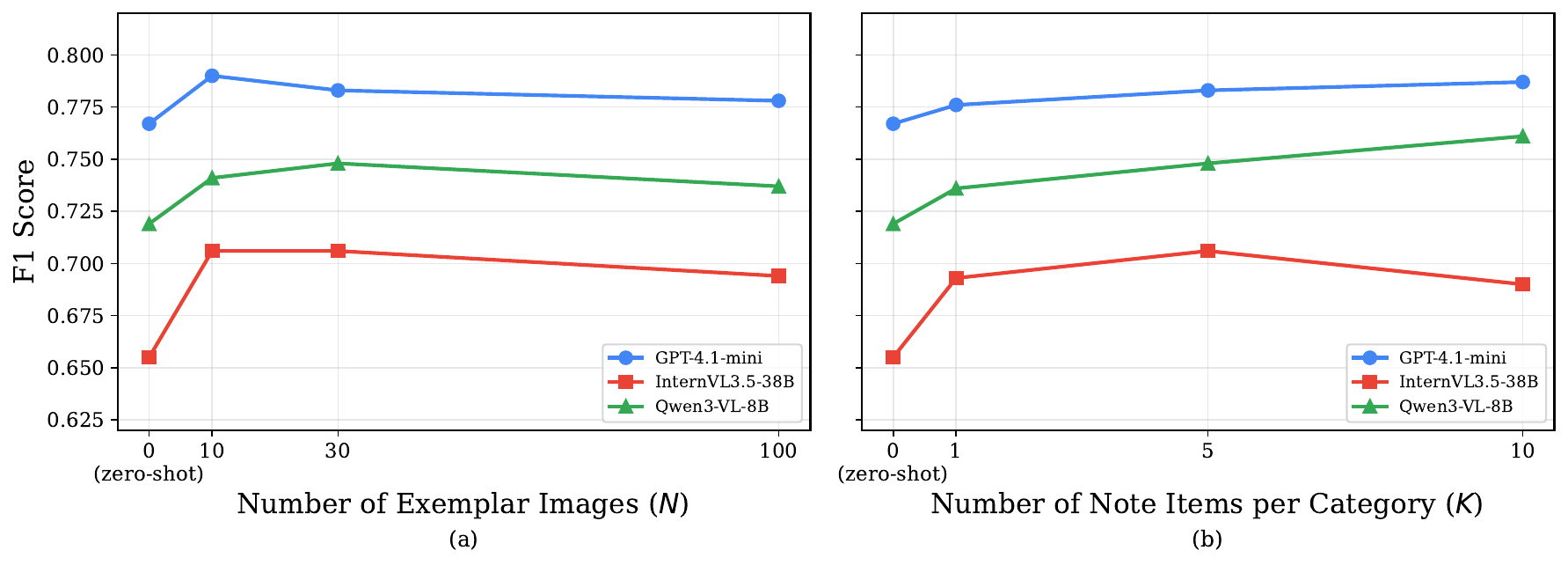}
\caption{
Ablation on note construction parameters.
\textbf{(a)}~F1 vs.\ the number of exemplar images $N$. Performance saturates at $N{\approx}30$, indicating that systematic error patterns can be surfaced from a modest exemplar set.
\textbf{(b)}~F1 vs.\ the maximum number of note items per category $K$. Even $K{=}1$ already yields strong gains, with performance improving slightly further at $K{=}5$.
}
\label{fig:ablation_scaling}
\end{figure*}

\noindent\textbf{Number of Exemplars and Note Items.}
Figure~\ref{fig:ablation_scaling} analyzes two key parameters of the offline phase using 100 images sampled from IIW-400 for efficient evaluation.

For the number of exemplar images $N$ (Figure~\ref{fig:ablation_scaling}(a)), all models show a substantial improvement from zero-shot to $N=10$, with performance largely plateauing around $N=30$ and slightly declining at $N=100$. Qualitative analysis of GPT-4.1-mini suggests that larger exemplar pools shift reflection notes from model-specific guidance toward generic instructions. For example, notes generated with $N=30$ contain targeted rules such as \textit{``Do not add unsupported details to signs, logos, or symbols,''} whereas those from $N=100$ become broader directives like \textit{``Avoid subjective or interpretive descriptions not clearly supported by the image or reference.''} This indicates that systematic error patterns can be reliably surfaced from a modest exemplar set, while larger pools introduce variation that dilutes the corrective signal.

For the maximum items per category $K$ (Figure~\ref{fig:ablation_scaling}(b)), even $K{=}1$ provides a substantial performance gain. Under this setting, models tend to produce a single reflection note that aggregates multiple corrective signals rather than a narrowly scoped rule. For example, a reflection note generated at $K{=}1$ states: \textit{``Include precise visible details of object features, spatial distributions, lighting and shadow effects, background elements, and compositional angles to ensure completeness and accuracy.''}
As $K$ increases (e.g., $K{=}5$), these aggregated instructions are decomposed into multiple more specialized reflection notes that introduce more specific guidelines, which further improves performance over the $K{=}1$ setting. While some models continue improving up to $K{=}10$, others peak around $K{=}5$ and slightly decline thereafter, suggesting that the optimal number of guidelines varies across models and may depend on their ability to incorporate multiple instructions.

\subsection{Case Study}
\label{sec:case_study}

\begin{figure*}[t]
\centering
\includegraphics[width=0.92\textwidth]{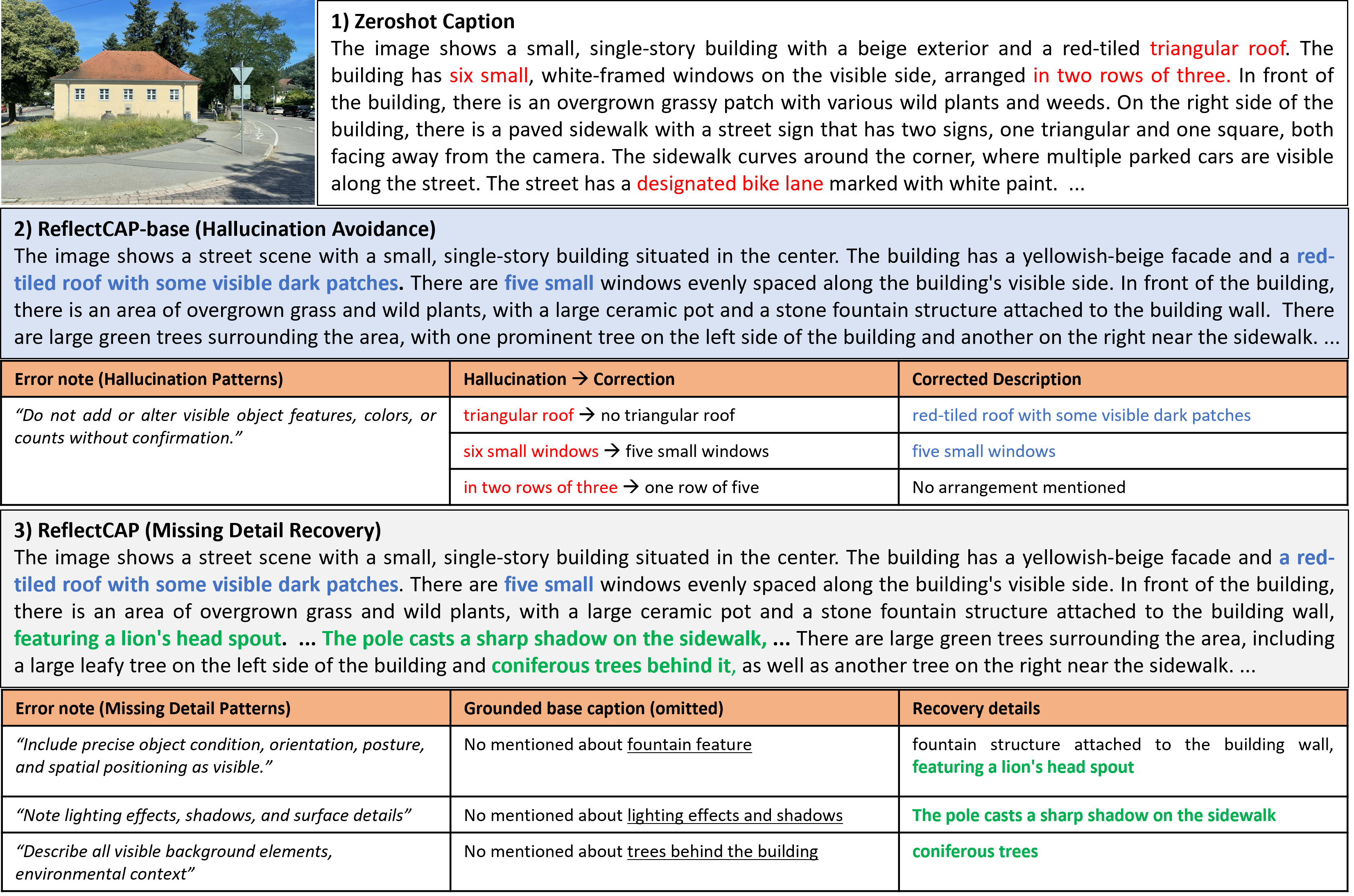}
\caption{
Case study of our pipeline.
\textbf{Top}: Zero-shot Caption
\textbf{Middle}: ReflectCAP-Base suppresses hallucinations via Avoid notes.
\textbf{Bottom}: ReflectCAP-Full recovers embossed text details guided by Include notes.
\textcolor{red}{Red} denotes hallucinated expressions, \textcolor{blue}{blue} denotes hallucination-corrected descriptions, and \textcolor{green}{green} denotes recovered fine-grained details.
}
\label{fig:case_study}
\end{figure*}

Figure~\ref{fig:case_study} illustrates a case study examining each pipeline stage of ReflectCAP using GPT-4.1 mini. In the zero-shot caption, hallucinations are observed in fine-grained details such as roof shape, window count and arrangement, and sign appearance. By applying our hallucination avoidance patterns, ReflectCAP corrects the window count from six to five and omits unverifiable arrangements, yielding a more factually grounded description. Furthermore, the missing detail recovery patterns enable the model to capture previously overlooked visual elements, including the lion's head fountain spout and cast shadows. These results demonstrate that ReflectCAP effectively encodes common error patterns into structured reflection notes and leverages them to steer caption generation toward greater accuracy and visual fidelity.

\section{Conclusion}
\label{sec:conclusion}

We presented ReflectCAP, a tuning-free framework that distills a target LVLM's recurring hallucination and omission patterns into Structured Reflection Notes. By leveraging these notes at caption generation time, ReflectCAP steers the model separately for each pattern type---suppressing hallucinations and recovering missing details---then merges the resulting captions into a single, comprehensive description. Across 8 LVLMs, ReflectCAP consistently achieves the highest F1 score on factuality--coverage evaluation, and substantially outperforms all baselines on CapArena-Auto. Furthermore, from a compute-efficiency perspective, ReflectCAP is more effective than both scaling up model size and scaling inference-time computation for improving caption quality.

\section*{Acknowledgements}
This work was partly supported by the National Research Foundation of Korea (NRF) grant funded by the Korea government (RS-2024-00348233), Institute of Information \& communications Technology Planning \& Evaluation (IITP) grant funded by the Korea government(MSIT) [NO.RS-2021-II211343, Artificial Intelligence Graduate School Program (Seoul National University)] \& RS-2021-II212068, Artificial Intelligence Innovation Hub (Artificial Intelligence Institute, Seoul National University)], and the BK21 FOUR program of the Education and Research Program for Future ICT Pioneers, Seoul National University in 2024.
K. Jung is with ASRI, Seoul National University, Korea.
\clearpage


%
%
\bibliographystyle{splncs04}
\bibliography{main}

@inproceedings{yanuka-etal-2025-bridging,
    title = "Bridging the Visual Gap: Fine-Tuning Multimodal Models with Knowledge-Adapted Captions",
    author = "Yanuka, Moran  and
      Ben-Kish, Assaf  and
      Bitton, Yonatan  and
      Szpektor, Idan  and
      Giryes, Raja",
    editor = "Chiruzzo, Luis  and
      Ritter, Alan  and
      Wang, Lu",
    booktitle = "Proceedings of the 2025 Conference of the Nations of the Americas Chapter of the Association for Computational Linguistics: Human Language Technologies (Volume 1: Long Papers)",
    month = apr,
    year = "2025",
    address = "Albuquerque, New Mexico",
    publisher = "Association for Computational Linguistics",
    url = "https://aclanthology.org/2025.naacl-long.527/",
    doi = "10.18653/v1/2025.naacl-long.527",
    pages = "10497--10518",
    ISBN = "979-8-89176-189-6"
}

@article{liu2023mitigating,
  title={Mitigating hallucination in large multi-modal models via robust instruction tuning},
  author={Liu, Fuxiao and Lin, Kevin and Li, Linjie and Wang, Jianfeng and Yacoob, Yaser and Wang, Lijuan},
  journal={arXiv preprint arXiv:2306.14565},
  year={2023}
}

@inproceedings{liu2024improved,
  title={Improved baselines with visual instruction tuning},
  author={Liu, Haotian and Li, Chunyuan and Li, Yuheng and Lee, Yong Jae},
  booktitle={Proceedings of the IEEE/CVF conference on computer vision and pattern recognition},
  pages={26296--26306},
  year={2024}
}

@article{dai2023instructblip,
  title={Instructblip: Towards general-purpose vision-language models with instruction tuning},
  author={Dai, Wenliang and Li, Junnan and Li, Dongxu and Tiong, Anthony and Zhao, Junqi and Wang, Weisheng and Li, Boyang and Fung, Pascale N and Hoi, Steven},
  journal={Advances in neural information processing systems},
  volume={36},
  pages={49250--49267},
  year={2023}
}

@inproceedings{favero2024multi,
  title={Multi-modal hallucination control by visual information grounding},
  author={Favero, Alessandro and Zancato, Luca and Trager, Matthew and Choudhary, Siddharth and Perera, Pramuditha and Achille, Alessandro and Swaminathan, Ashwin and Soatto, Stefano},
  booktitle={Proceedings of the IEEE/CVF Conference on Computer Vision and Pattern Recognition},
  pages={14303--14312},
  year={2024}
}

@misc{chung2026v1learningpointvisual,
      title={v1: Learning to Point Visual Tokens for Multimodal Grounded Reasoning}, 
      author={Jiwan Chung and Junhyeok Kim and Siyeol Kim and Jaeyoung Lee and Min Soo Kim and Youngjae Yu},
      year={2026},
      eprint={2505.18842},
      archivePrefix={arXiv},
      primaryClass={cs.CL},
      url={https://arxiv.org/abs/2505.18842}, 
}

@inproceedings{zhu2025ibd,
  title={Ibd: Alleviating hallucinations in large vision-language models via image-biased decoding},
  author={Zhu, Lanyun and Ji, Deyi and Chen, Tianrun and Xu, Peng and Ye, Jieping and Liu, Jun},
  booktitle={Proceedings of the Computer Vision and Pattern Recognition Conference},
  pages={1624--1633},
  year={2025}
}

@inproceedings{wang2024mitigating,
  title={Mitigating hallucinations in large vision-language models with instruction contrastive decoding},
  author={Wang, Xintong and Pan, Jingheng and Ding, Liang and Biemann, Chris},
  booktitle={Findings of the Association for Computational Linguistics: ACL 2024},
  pages={15840--15853},
  year={2024}
}

@misc{ouyang2025reasoningbankscalingagentselfevolving,
      title={ReasoningBank: Scaling Agent Self-Evolving with Reasoning Memory}, 
      author={Siru Ouyang and Jun Yan and I-Hung Hsu and Yanfei Chen and Ke Jiang and Zifeng Wang and Rujun Han and Long T. Le and Samira Daruki and Xiangru Tang and Vishy Tirumalashetty and George Lee and Mahsan Rofouei and Hangfei Lin and Jiawei Han and Chen-Yu Lee and Tomas Pfister},
      year={2025},
      eprint={2509.25140},
      archivePrefix={arXiv},
      primaryClass={cs.AI},
      url={https://arxiv.org/abs/2509.25140}, 
}

@inproceedings{zhang2025sc,
  title={Sc-captioner: Improving image captioning with self-correction by reinforcement learning},
  author={Zhang, Lin and Zeng, Xianfang and Li, Kangcong and Yu, Gang and Chen, Tao},
  booktitle={Proceedings of the IEEE/CVF International Conference on Computer Vision},
  pages={23145--23155},
  year={2025}
}

@inproceedings{he2025self,
  title={Self-correction is more than refinement: A learning framework for visual and language reasoning tasks},
  author={He, Jiayi and Lin, Hehai and Wang, Qingyun and Fung, Yi R and Ji, Heng},
  booktitle={Findings of the Association for Computational Linguistics: ACL 2025},
  pages={6405--6421},
  year={2025}
}

@inproceedings{yue-etal-2024-less,
    title = "Less is More: Mitigating Multimodal Hallucination from an {EOS} Decision Perspective",
    author = "Yue, Zihao  and
      Zhang, Liang  and
      Jin, Qin",
    editor = "Ku, Lun-Wei  and
      Martins, Andre  and
      Srikumar, Vivek",
    booktitle = "Proceedings of the 62nd Annual Meeting of the Association for Computational Linguistics (Volume 1: Long Papers)",
    month = aug,
    year = "2024",
    address = "Bangkok, Thailand",
    publisher = "Association for Computational Linguistics",
    url = "https://aclanthology.org/2024.acl-long.633/",
    doi = "10.18653/v1/2024.acl-long.633",
    pages = "11766--11781"
}

@article{kaplan2020scaling,
  title={Scaling laws for neural language models},
  author={Kaplan, Jared and McCandlish, Sam and Henighan, Tom and Brown, Tom B and Chess, Benjamin and Child, Rewon and Gray, Scott and Radford, Alec and Wu, Jeffrey and Amodei, Dario},
  journal={arXiv preprint arXiv:2001.08361},
  year={2020}
}

@inproceedings{cheng-etal-2025-caparena,
    title = "{C}ap{A}rena: Benchmarking and Analyzing Detailed Image Captioning in the {LLM} Era",
    author = "Cheng, Kanzhi  and
      Song, Wenpo  and
      Fan, Jiaxin  and
      Ma, Zheng  and
      Sun, Qiushi  and
      Xu, Fangzhi  and
      Yan, Chenyang  and
      Chen, Nuo  and
      Zhang, Jianbing  and
      Chen, Jiajun",
    editor = "Che, Wanxiang  and
      Nabende, Joyce  and
      Shutova, Ekaterina  and
      Pilehvar, Mohammad Taher",
    booktitle = "Findings of the Association for Computational Linguistics: ACL 2025",
    month = jul,
    year = "2025",
    address = "Vienna, Austria",
    publisher = "Association for Computational Linguistics",
    url = "https://aclanthology.org/2025.findings-acl.724/",
    doi = "10.18653/v1/2025.findings-acl.724",
    pages = "14077--14094",
    ISBN = "979-8-89176-256-5"
}

@article{marsili2025same,
  title={Same or Not? Enhancing Visual Perception in Vision-Language Models},
  author={Marsili, Damiano and Mehta, Aditya and Lin, Ryan Y and Gkioxari, Georgia},
  journal={arXiv preprint arXiv:2512.23592},
  year={2025}
}

@misc{onoe2024doccidescriptionsconnectedcontrasting,
      title={DOCCI: Descriptions of Connected and Contrasting Images}, 
      author={Yasumasa Onoe and Sunayana Rane and Zachary Berger and Yonatan Bitton and Jaemin Cho and Roopal Garg and Alexander Ku and Zarana Parekh and Jordi Pont-Tuset and Garrett Tanzer and Su Wang and Jason Baldridge},
      year={2024},
      eprint={2404.19753},
      archivePrefix={arXiv},
      primaryClass={cs.CV},
      url={https://arxiv.org/abs/2404.19753}, 
}

@article{rahmanzadehgervi2024vision,
  title={Vision language models are blind: Failing to translate detailed visual features into words},
  author={Rahmanzadehgervi, Pooyan and Bolton, Logan and Taesiri, Mohammad Reza and Nguyen, Anh Totti},
  journal={arXiv preprint arXiv:2407.06581},
  year={2024}
}

@inproceedings{fu2024blink,
  title={Blink: Multimodal large language models can see but not perceive},
  author={Fu, Xingyu and Hu, Yushi and Li, Bangzheng and Feng, Yu and Wang, Haoyu and Lin, Xudong and Roth, Dan and Smith, Noah A and Ma, Wei-Chiu and Krishna, Ranjay},
  booktitle={European Conference on Computer Vision},
  pages={148--166},
  year={2024},
  organization={Springer}
}

@article{betker2023improving,
  title={Improving image generation with better captions},
  author={Betker, James and Goh, Gabriel and Jing, Li and Brooks, Tim and Wang, Jianfeng and Li, Linjie and Ouyang, Long and Zhuang, Juntang and Lee, Joyce and Guo, Yufei and others},
  journal={Computer Science. https://cdn. openai. com/papers/dall-e-3. pdf},
  volume={2},
  number={3},
  pages={8},
  year={2023}
}

@article{gutflaish2025generating,
  title={Generating an Image From 1,000 Words: Enhancing Text-to-Image With Structured Captions},
  author={Gutflaish, Eyal and Kachlon, Eliran and Zisman, Hezi and Hacham, Tal and Sarid, Nimrod and Visheratin, Alexander and Huberman, Saar and Davidi, Gal and Bukchin, Guy and Goldberg, Kfir and others},
  journal={arXiv preprint arXiv:2511.06876},
  year={2025}
}

@misc{merchant2025structuredcaptionsimproveprompt,
      title={Structured Captions Improve Prompt Adherence in Text-to-Image Models (Re-LAION-Caption 19M)}, 
      author={Nicholas Merchant and Haitz Sáez de Ocáriz Borde and Andrei Cristian Popescu and Carlos Garcia Jurado Suarez},
      year={2025},
      eprint={2507.05300},
      archivePrefix={arXiv},
      primaryClass={cs.CV},
      url={https://arxiv.org/abs/2507.05300}, 
}

@article{brooks2024video,
  title={Video generation models as world simulators},
  author={Brooks, Tim and Peebles, Bill and Holmes, Connor and DePue, Will and Guo, Yufei and Jing, Leo and Schnurr, David and Taylor, Joe and Luhman, Troy and Luhman, Eric and others},
  journal={OpenAI Blog},
  volume={1},
  number={8},
  pages={1},
  year={2024}
}

@article{ju2024miradata,
  title={Miradata: A large-scale video dataset with long durations and structured captions},
  author={Ju, Xuan and Gao, Yiming and Zhang, Zhaoyang and Yuan, Ziyang and Wang, Xintao and Zeng, Ailing and Xiong, Yu and Xu, Qiang and Shan, Ying},
  journal={Advances in Neural Information Processing Systems},
  volume={37},
  pages={48955--48970},
  year={2024}
}

@inproceedings{garg-etal-2024-imageinwords,

    title = "{I}mage{I}n{W}ords: Unlocking Hyper-Detailed Image Descriptions",

    author = "Garg, Roopal  and

      Burns, Andrea  and

      Karagol Ayan, Burcu  and

      Bitton, Yonatan  and

      Montgomery, Ceslee  and

      Onoe, Yasumasa  and

      Bunner, Andrew  and

      Krishna, Ranjay  and

      Baldridge, Jason Michael  and

      Soricut, Radu",

    editor = "Al-Onaizan, Yaser  and

      Bansal, Mohit  and

      Chen, Yun-Nung",

    booktitle = "Proceedings of the 2024 Conference on Empirical Methods in Natural Language Processing",

    month = nov,

    year = "2024",

    address = "Miami, Florida, USA",

    publisher = "Association for Computational Linguistics",

    url = "https://aclanthology.org/2024.emnlp-main.6/",

    doi = "10.18653/v1/2024.emnlp-main.6",

    pages = "93--127"

}

@inproceedings{sun-etal-2025-mitigating-visual,
    title = "Mitigating Visual Forgetting via Take-along Visual Conditioning for Multi-modal Long {C}o{T} Reasoning",
    author = "Sun, Hai-Long  and
      Sun, Zhun  and
      Peng, Houwen  and
      Ye, Han-Jia",
    editor = "Che, Wanxiang  and
      Nabende, Joyce  and
      Shutova, Ekaterina  and
      Pilehvar, Mohammad Taher",
    booktitle = "Proceedings of the 63rd Annual Meeting of the Association for Computational Linguistics (Volume 1: Long Papers)",
    month = jul,
    year = "2025",
    address = "Vienna, Austria",
    publisher = "Association for Computational Linguistics",
    url = "https://aclanthology.org/2025.acl-long.257/",
    doi = "10.18653/v1/2025.acl-long.257",
    pages = "5158--5171",
    ISBN = "979-8-89176-251-0"
}

@inproceedings{
li2025the,
title={The Hidden Life of Tokens: Reducing Hallucination of Large Vision-Language Models Via Visual Information Steering},
author={Zhuowei Li and Haizhou Shi and Yunhe Gao and Di Liu and Zhenting Wang and Yuxiao Chen and Ting Liu and Long Zhao and Hao Wang and Dimitris N. Metaxas},
booktitle={Forty-second International Conference on Machine Learning},
year={2025},
url={https://openreview.net/forum?id=7BKcLeHQsm}
}

@misc{wan2025compassenhancingagentlonghorizon,
      title={COMPASS: Enhancing Agent Long-Horizon Reasoning with Evolving Context}, 
      author={Guangya Wan and Mingyang Ling and Xiaoqi Ren and Rujun Han and Sheng Li and Zizhao Zhang},
      year={2025},
      eprint={2510.08790},
      archivePrefix={arXiv},
      primaryClass={cs.AI},
      url={https://arxiv.org/abs/2510.08790}, 
}

@article{zhu2026toward,
  title={Toward ultra-long-horizon agentic science: Cognitive accumulation for machine learning engineering},
  author={Zhu, Xinyu and Cai, Yuzhu and Liu, Zexi and Zheng, Bingyang and Wang, Cheng and Ye, Rui and Chen, Jiaao and Wang, Hanrui and Wang, Wei-Chen and Zhang, Yuzhi and others},
  journal={arXiv preprint arXiv:2601.10402},
  year={2026}
}

@inproceedings{tan2025prospect,
  title={In prospect and retrospect: Reflective memory management for long-term personalized dialogue agents},
  author={Tan, Zhen and Yan, Jun and Hsu, I-Hung and Han, Rujun and Wang, Zifeng and Le, Long and Song, Yiwen and Chen, Yanfei and Palangi, Hamid and Lee, George and others},
  booktitle={Proceedings of the 63rd Annual Meeting of the Association for Computational Linguistics (Volume 1: Long Papers)},
  pages={8416--8439},
  year={2025}
}

@misc{madaan2023selfrefineiterativerefinementselffeedback,
      title={Self-Refine: Iterative Refinement with Self-Feedback}, 
      author={Aman Madaan and Niket Tandon and Prakhar Gupta and Skyler Hallinan and Luyu Gao and Sarah Wiegreffe and Uri Alon and Nouha Dziri and Shrimai Prabhumoye and Yiming Yang and Shashank Gupta and Bodhisattwa Prasad Majumder and Katherine Hermann and Sean Welleck and Amir Yazdanbakhsh and Peter Clark},
      year={2023},
      eprint={2303.17651},
      archivePrefix={arXiv},
      primaryClass={cs.CL},
      url={https://arxiv.org/abs/2303.17651}, 
}

@article{shinn2023reflexion,
  title={Reflexion: Language agents with verbal reinforcement learning},
  author={Shinn, Noah and Cassano, Federico and Gopinath, Ashwin and Narasimhan, Karthik and Yao, Shunyu},
  journal={Advances in neural information processing systems},
  volume={36},
  pages={8634--8652},
  year={2023}
}

@inproceedings{anderson2016spice,
  title={Spice: Semantic propositional image caption evaluation},
  author={Anderson, Peter and Fernando, Basura and Johnson, Mark and Gould, Stephen},
  booktitle={European conference on computer vision},
  pages={382--398},
  year={2016},
  organization={Springer}
}

@inproceedings{banerjee2005meteor,
  title={METEOR: An automatic metric for MT evaluation with improved correlation with human judgments},
  author={Banerjee, Satanjeev and Lavie, Alon},
  booktitle={Proceedings of the acl workshop on intrinsic and extrinsic evaluation measures for machine translation and/or summarization},
  pages={65--72},
  year={2005}
}

@inproceedings{papineni2002bleu,
  title={Bleu: a method for automatic evaluation of machine translation},
  author={Papineni, Kishore and Roukos, Salim and Ward, Todd and Zhu, Wei-Jing},
  booktitle={Proceedings of the 40th annual meeting of the Association for Computational Linguistics},
  pages={311--318},
  year={2002}
}

@article{kamoi-etal-2024-llms,
    title = "When Can {LLM}s Actually Correct Their Own Mistakes? A Critical Survey of Self-Correction of {LLM}s",
    author = "Kamoi, Ryo  and
      Zhang, Yusen  and
      Zhang, Nan  and
      Han, Jiawei  and
      Zhang, Rui",
    journal = "Transactions of the Association for Computational Linguistics",
    volume = "12",
    year = "2024",
    address = "Cambridge, MA",
    publisher = "MIT Press",
    url = "https://aclanthology.org/2024.tacl-1.78/",
    doi = "10.1162/tacl_a_00713",
    pages = "1417--1440"
}

@inproceedings{huang2024opera,
  title={Opera: Alleviating hallucination in multi-modal large language models via over-trust penalty and retrospection-allocation},
  author={Huang, Qidong and Dong, Xiaoyi and Zhang, Pan and Wang, Bin and He, Conghui and Wang, Jiaqi and Lin, Dahua and Zhang, Weiming and Yu, Nenghai},
  booktitle={Proceedings of the IEEE/CVF Conference on Computer Vision and Pattern Recognition},
  pages={13418--13427},
  year={2024}
}

@inproceedings{leng2024mitigating,
  title={Mitigating object hallucinations in large vision-language models through visual contrastive decoding},
  author={Leng, Sicong and Zhang, Hang and Chen, Guanzheng and Li, Xin and Lu, Shijian and Miao, Chunyan and Bing, Lidong},
  booktitle={Proceedings of the IEEE/CVF Conference on Computer Vision and Pattern Recognition},
  pages={13872--13882},
  year={2024}
}

@article{zhou2023analyzing,
  title={Analyzing and mitigating object hallucination in large vision-language models},
  author={Zhou, Yiyang and Cui, Chenhang and Yoon, Jaehong and Zhang, Linjun and Deng, Zhun and Finn, Chelsea and Bansal, Mohit and Yao, Huaxiu},
  journal={arXiv preprint arXiv:2310.00754},
  year={2023}
}

@article{liu2023visual,
  title={Visual instruction tuning},
  author={Liu, Haotian and Li, Chunyuan and Wu, Qingyang and Lee, Yong Jae},
  journal={Advances in neural information processing systems},
  volume={36},
  pages={34892--34916},
  year={2023}
}

@article{zhu2023minigpt,
  title={Minigpt-4: Enhancing vision-language understanding with advanced large language models},
  author={Zhu, Deyao and Chen, Jun and Shen, Xiaoqian and Li, Xiang and Elhoseiny, Mohamed},
  journal={arXiv preprint arXiv:2304.10592},
  year={2023}
}

@inproceedings{min-etal-2025-mitigating,
    title = "Mitigating Hallucinations in Large Vision-Language Models via Summary-Guided Decoding",
    author = "Min, Kyungmin  and
      Kim, Minbeom  and
      Lee, Kang-il  and
      Lee, Dongryeol  and
      Jung, Kyomin",
    editor = "Chiruzzo, Luis  and
      Ritter, Alan  and
      Wang, Lu",
    booktitle = "Findings of the Association for Computational Linguistics: NAACL 2025",
    month = apr,
    year = "2025",
    address = "Albuquerque, New Mexico",
    publisher = "Association for Computational Linguistics",
    url = "https://aclanthology.org/2025.findings-naacl.235/",
    doi = "10.18653/v1/2025.findings-naacl.235",
    pages = "4183--4198",
    ISBN = "979-8-89176-195-7"
}

@inproceedings{
lee2025toward,
title={Toward Robust Hyper-Detailed Image Captioning: A Multiagent Approach and Dual Evaluation Metrics for Factuality and Coverage},
author={Saehyung Lee and Seunghyun Yoon and Trung Bui and Jing Shi and Sungroh Yoon},
booktitle={Forty-second International Conference on Machine Learning},
year={2025},
url={https://openreview.net/forum?id=REnIf3dCsI}
}

@inproceedings{lee-etal-2025-vlind,
    title = "{VL}ind-Bench: Measuring Language Priors in Large Vision-Language Models",
    author = "Lee, Kang-il  and
      Kim, Minbeom  and
      Yoon, Seunghyun  and
      Kim, Minsung  and
      Lee, Dongryeol  and
      Koh, Hyukhun  and
      Jung, Kyomin",
    editor = "Chiruzzo, Luis  and
      Ritter, Alan  and
      Wang, Lu",
    booktitle = "Findings of the Association for Computational Linguistics: NAACL 2025",
    month = apr,
    year = "2025",
    address = "Albuquerque, New Mexico",
    publisher = "Association for Computational Linguistics",
    url = "https://aclanthology.org/2025.findings-naacl.231/",
    doi = "10.18653/v1/2025.findings-naacl.231",
    pages = "4129--4144",
    ISBN = "979-8-89176-195-7"
}

@misc{urbanek2024pictureworth77text,
      title={A Picture is Worth More Than 77 Text Tokens: Evaluating CLIP-Style Models on Dense Captions}, 
      author={Jack Urbanek and Florian Bordes and Pietro Astolfi and Mary Williamson and Vasu Sharma and Adriana Romero-Soriano},
      year={2024},
      eprint={2312.08578},
      archivePrefix={arXiv},
      primaryClass={cs.CV},
      url={https://arxiv.org/abs/2312.08578}, 
}

@inproceedings{zheng2024llamafactory,
  title={Llamafactory: Unified efficient fine-tuning of 100+ language models},
  author={Zheng, Yaowei and Zhang, Richong and Zhang, Junhao and Ye, Yanhan and Luo, Zheyan},
  booktitle={Proceedings of the 62nd annual meeting of the association for computational linguistics (volume 3: system demonstrations)},
  pages={400--410},
  year={2024}
}

@misc{rostamzadeh2018fashiongengenerativefashiondataset,
      title={Fashion-Gen: The Generative Fashion Dataset and Challenge}, 
      author={Negar Rostamzadeh and Seyedarian Hosseini and Thomas Boquet and Wojciech Stokowiec and Ying Zhang and Christian Jauvin and Chris Pal},
      year={2018},
      eprint={1806.08317},
      archivePrefix={arXiv},
      primaryClass={stat.ML},
      url={https://arxiv.org/abs/1806.08317}, 
}

@misc{yang2026captionqacaptionusefulimage,
      title={CaptionQA: Is Your Caption as Useful as the Image Itself?}, 
      author={Shijia Yang and Yunong Liu and Bohan Zhai and Ximeng Sun and Zicheng Liu and Emad Barsoum and Manling Li and Chenfeng Xu},
      year={2026},
      eprint={2511.21025},
      archivePrefix={arXiv},
      primaryClass={cs.CV},
      url={https://arxiv.org/abs/2511.21025}, 
}

@article{liu2026capability,
  title={Capability: A comprehensive visual caption benchmark for evaluating both correctness and thoroughness},
  author={Liu, Zhihang and Xie, Chen-Wei and Wen, Bin and Yu, Feiwu and Li, Pandeng and Zhang, Boqiang and Yang, Nianzu and Gao, Zuan and Zheng, Yun and Xie, Hongtao},
  journal={Advances in Neural Information Processing Systems},
  volume={38},
  year={2026}
}

@article{chen2015microsoft,
  title={Microsoft coco captions: Data collection and evaluation server},
  author={Chen, Xinlei and Fang, Hao and Lin, Tsung-Yi and Vedantam, Ramakrishna and Gupta, Saurabh and Doll{\'a}r, Piotr and Zitnick, C Lawrence},
  journal={arXiv preprint arXiv:1504.00325},
  year={2015}
}

@inproceedings{lin2004rouge,
  title={Rouge: A package for automatic evaluation of summaries},
  author={Lin, Chin-Yew},
  booktitle={Text summarization branches out},
  pages={74--81},
  year={2004}
}

@inproceedings{vedantam2015cider,
  title={Cider: Consensus-based image description evaluation},
  author={Vedantam, Ramakrishna and Lawrence Zitnick, C and Parikh, Devi},
  booktitle={Proceedings of the IEEE conference on computer vision and pattern recognition},
  pages={4566--4575},
  year={2015}
}

@article{xing2025caprl,
  title={Caprl: Stimulating dense image caption capabilities via reinforcement learning},
  author={Xing, Long and Dong, Xiaoyi and Zang, Yuhang and Cao, Yuhang and Liang, Jianze and Huang, Qidong and Wang, Jiaqi and Wu, Feng and Lin, Dahua},
  journal={arXiv preprint arXiv:2509.22647},
  year={2025}
}
%
\tcbset{
  promptbox/.style={
    enhanced, breakable,
    colback=gray!5, colframe=gray!40,
    left=8pt, right=8pt, top=6pt, bottom=6pt,
    fontupper=\small,
  },
}

\appendix
\section*{Appendix}

\section{Supervised Fine-tuning Details}
\label{sec:sft_detail}
We fine-tune InternVL3.5-4B and Qwen2.5-VL-7B on captions corresponding to the DOCCI images. The captions come from two sources: the original human-authored captions provided in DOCCI and captions generated by our ReflectCAP pipeline.
We apply LoRA (rank $= 64$, $\alpha = 128$, dropout $= 0.05$) to all linear layers of the language model.
Training runs for 3 epochs with a batch size of 2 per device and 4 gradient accumulation steps, using a learning rate of $1 \times 10^{-4}$ with cosine scheduling and a 3\% warmup ratio.
All models are trained in BF16 precision on two NVIDIA A6000 GPUs using the LLaMA-Factory framework~\cite{zheng2024llamafactory}.
For consistency, we report results from the final checkpoint after 3 training epochs for all supervised fine-tuning experiments.
Per-epoch evaluation results are reported in Table~\ref{tab:per_epoch}; evaluation follows the same protocol as in Section~\ref{sec:experiments}.

\begin{table}[htbp]
\centering
\small
\setlength{\tabcolsep}{7pt}
\renewcommand{\arraystretch}{1.1}
\caption{Per-epoch evaluation results for fine-tuned models on ReflectCAP-generated and human-authored captions.}
\label{tab:per_epoch}

\begin{adjustbox}{width=0.8\textwidth}
\begin{tabular}{llcrrr}
\toprule
\textbf{Model} & \textbf{Train Data} & \textbf{Epoch} & \textbf{Precision} & \textbf{Recall} & \textbf{F1} \\
\midrule
\multirow{6}{*}{Qwen2.5-VL-7B}
  & \multirow{3}{*}{ReflectCAP}
    & 1 & 0.7112 & 0.6228 & 0.6674 \\
  & & 2 & 0.7065 & 0.6284 & 0.6651 \\
  & & 3 & 0.6993 & 0.6346 & 0.6654 \\
\cmidrule{2-6}
  & \multirow{3}{*}{Human-authored}
    & 1 & 0.5702 & 0.5567 & 0.5634 \\
  & & 2 & 0.5689 & 0.5754 & 0.5721 \\
  & & 3 & 0.5715 & 0.5798 & 0.5756 \\
\midrule
\addlinespace[2pt]
\multirow{6}{*}{InternVL3.5-4B}
  & \multirow{3}{*}{ReflectCAP}
    & 1 & 0.6885 & 0.6097 & 0.6467 \\
  & & 2 & 0.6943 & 0.6085 & 0.6486 \\
  & & 3 & 0.6649 & 0.6157 & 0.6393 \\
\cmidrule{2-6}
  & \multirow{3}{*}{Human-authored}
    & 1 & 0.5940 & 0.5602 & 0.5766 \\
  & & 2 & 0.5763 & 0.5669 & 0.5716 \\
  & & 3 & 0.5821 & 0.5653 & 0.5736 \\
\addlinespace[2pt]
\bottomrule
\end{tabular}
\end{adjustbox}

\end{table}

\section{Evaluation with Conventional Captioning Metrics}
\label{sec}

Although conventional n-gram and overlap-based captioning metrics are less suited to long-form detailed captioning, we additionally report them as supplementary evidence.
Short COCO-style\cite{chen2015microsoft} references often provide limited signal for paragraph-length captions and do not directly measure factuality or coverage.
Therefore, we compute BLEU-4\cite{papineni2002bleu}, ROUGE-L\cite{lin2004rouge}, METEOR\cite{banerjee2005meteor}, CIDEr\cite{vedantam2015cider}, and SPICE\cite{anderson2016spice} on IIW-400, whose long reference captions are more aligned with detailed image captioning.

As shown in Table~\ref{tab:traditional}, ReflectCAP generally improves conventional captioning metrics over zero-shot and CapMAS baselines.
On GPT-4.1-mini, ReflectCAP improves all reported metrics.
On Qwen3-VL-8B, ReflectCAP improves BLEU-4, ROUGE-L, METEOR, and SPICE, while CIDEr does not improve.
We therefore treat these metrics as complementary to the factuality and coverage-oriented evaluation used in the main experiments.

\begin{table}[!h]
\centering
\caption{\textbf{Results of traditional captioning metrics}. B-4/R-L/M/C/S = BLEU-4/ROUGE-L/METEOR/CIDEr/SPICE.}
\label{tab:traditional}

\resizebox{0.7\columnwidth}{!}{%
\begin{tabular}{llccccc}
\toprule
Backbone & Method & B-4 & R-L & M & C & S \\
\midrule
\multirow{3}{*}{GPT-4.1-mini}
 & ZS & 2.77 & 18.46 & 14.60 & 4.11 & 17.81 \\
 & CapMAS & 2.74 & 18.36 & 14.34 & 3.62 & 17.67 \\
 & \textbf{Ours} & \textbf{4.93} & \textbf{19.84} & \textbf{19.24} & \textbf{4.57} & \textbf{18.59} \\
\midrule
\multirow{3}{*}{Qwen3-VL-8B}
 & ZS & 2.80 & 17.17 & 15.78 & \textbf{3.99} & 16.75 \\
 & CapMAS & 2.68 & 17.14 & 15.46 & 3.91 & 16.71 \\
 & \textbf{Ours} & \textbf{3.64} & \textbf{17.91} & \textbf{18.42} & 3.19 & \textbf{17.80} \\
\bottomrule
\end{tabular}}
\end{table}

\section{Comparison with Contrastive Decoding and RL-Based Baselines}
\label{sec}
We also compare ReflectCAP with additional baseline methods. Specifically, we evaluate VCD\cite{leng2024mitigating}, a decoding-based hallucination mitigation method, and CapRL\cite{xing2025caprl}, an RL-based method for dense captioning. For VCD, we evaluate on a 100-image subset of IIW-400 using Qwen3-VL-8B. For CapRL, we compare against ReflectCAP on Qwen2.5-VL-7B under a matched image budget between CapRL training and ReflectCAP's offline note construction. Table~\ref{tab:decoding_rl_baselines} shows that VCD increases recall but substantially lowers precision, indicating a trade-off between coverage and factuality. ReflectCAP achieves the best F1 score by maintaining higher precision while preserving competitive recall. Under the matched-budget comparison with CapRL, ReflectCAP improves precision, recall, and F1. These results suggest that ReflectCAP remains competitive against both decoding-based hallucination mitigation and RL-based dense captioning baselines.

\begin{table}[!h]
\centering
\caption{Comparison with additional baselines.}
\label{tab:decoding_rl_baselines}
\resizebox{0.7\columnwidth}{!}{%
\begin{tabular}{llccc}
\toprule
Backbone & Method & Factuality & Coverage & F1 \\
\midrule
\multirow{2}{*}{Qwen3-VL-8B}
& VCD & 0.697 & \textbf{0.758} & 0.726 \\
& \textbf{Ours} & \textbf{0.776} & 0.724 & \textbf{0.749} \\
\midrule
\multirow{2}{*}{Qwen2.5-VL-7B}
& CapRL & 0.684 & 0.610 & 0.645 \\
& \textbf{Ours} & \textbf{0.696} & \textbf{0.637} & \textbf{0.665} \\
\bottomrule
\end{tabular}}
\end{table}

\section{Effect of the Grounded Base Caption} \label{sec:grounded}
We isolate the effect of the grounded base caption, which applies only the hallucination-avoidance note before the detail-focused captioning and merging stages. This ablation evaluates how much of ReflectCAP's improvement comes from suppressing common hallucination patterns alone.
As shown in Table~\ref{tab:grounded_base}, the grounded base caption improves factuality over zero-shot captioning on both GPT-4.1-mini and Qwen3-VL-8B. The full ReflectCAP pipeline achieves the best F1 on both models, indicating that hallucination avoidance alone is useful but insufficient; the detail-focused captioning and merging stages are needed to recover additional visual coverage.

\begin{table}[t]
\centering
\small
\setlength{\tabcolsep}{7pt}
\renewcommand{\arraystretch}{1.08}
\caption{Effect of using only the grounded base caption.}
\label{tab:grounded_base}
\begin{adjustbox}{width=0.78\textwidth}
\begin{tabular}{llccc}
\toprule
\textbf{Model} & \textbf{Condition} & \textbf{Factuality} & \textbf{Coverage} & \textbf{F1} \\
\midrule
\multirow{3}{*}{GPT-4.1-mini}
& Zero-shot & 83.6 & 68.1 & 75.1 \\
& Grounded Base Caption & \textbf{85.9} & 66.8 & 75.2 \\
& ReflectCAP (Full) & 83.8 & \textbf{72.0} & \textbf{77.5} \\
\midrule
\multirow{3}{*}{Qwen3-VL-8B}
& Zero-shot & 76.3 & 69.2 & 72.6 \\
& Grounded Base Caption & \textbf{77.4} & 69.5 & 73.2 \\
& ReflectCAP (Full) & 77.0 & \textbf{71.2} & \textbf{74.0} \\
\bottomrule
\end{tabular}
\end{adjustbox}
\end{table}

\section{Prompt Templates}
\label{sec:prompts}

We report here the full prompt templates used in the offline and online stages of our framework. The offline templates define the architectural roles of the three agents---the Captioning Agent, Feedback Agent, and Note Organizer---which collaborate to construct the Error Notes from training data. The online templates are used at inference time to generate grounded base captions, extract commonly missed details, and merge the two into a final refined caption.

\subsection{Offline Stage Prompts}
\label{sec:prompts-offline}

\noindent\textbf{Captioning Agent.}
\begin{tcolorbox}[promptbox]
\textbf{System:} You are an expert image captioner. Describe images accurately and in detail.

\medskip
\textbf{User:} Describe this image in detail.
\end{tcolorbox}

\noindent\textbf{Feedback Agent.}
\begin{tcolorbox}[promptbox]
\textbf{System:} You are a caption quality monitor. Compare the generated caption with the reference caption.

\smallskip
Your task:
\begin{enumerate}[leftmargin=1.8em, topsep=2pt, itemsep=1pt, parsep=0pt]
  \item Identify \textbf{HALLUCINATIONS}: details in the generated caption that are \textbf{WRONG} or \textbf{NOT visible} in the image.
  \item Identify \textbf{MISSING DETAILS}: important details in the reference caption that are \textbf{MISSING} from the generated caption.
\end{enumerate}
For each issue, provide: (1) what the issue is, (2) why it's problematic, (3) a simple rule to avoid/fix it.

\smallskip
Output format:
\begin{quote}\ttfamily\small
Hallucinations: - issue 1, - issue 2, \ldots\\
Missing Details: - issue 1, - issue 2, \ldots
\end{quote}
If no issues are found in a category, write ``None''.

\medskip
\textbf{User:}\\
\texttt{Generated Caption: \{generated\_caption\}}\\
\texttt{Reference Caption: \{reference\_caption\}}\\[4pt]
Analyze the generated caption against the reference and the image.
\end{tcolorbox}

\noindent\textbf{Note Organizer.}
\begin{tcolorbox}[promptbox]
\textbf{System:} You manage ``Error Notes'' for an image captioning model.

\smallskip
Your task:
\begin{enumerate}[leftmargin=1.8em, topsep=2pt, itemsep=1pt, parsep=0pt]
  \item Review new issues from this batch.
  \item Update the error notes by adding new issues, merging similar ones, summarizing into general rules, and keeping \textbf{maximum $k$ items} per category.
  \item Each item should be \textbf{simple and compact} (one line).
\end{enumerate}

\smallskip
Output format:
\begin{quote}\ttfamily\small
{[}Hallucination - Avoid These{]}: - item 1, - item 2, \ldots\ (max $k$)\\
{[}Missing Detail - Include These{]}: - item 1, - item 2, \ldots\ (max $k$)
\end{quote}

\medskip
\textbf{User:}\\
\texttt{Current Error Notes: \{current\_notes\}}\\
\texttt{New Issues from Batch: \{batch\_issues\}}\\[4pt]
Update the Error Notes. Keep it compact (max $k$ items per category).
\end{tcolorbox}

\subsection{Online Stage Prompts}
\label{sec:prompts-online}

\noindent\textbf{Stage 1: Grounded Base Caption.}
\begin{tcolorbox}[promptbox]
\textbf{System:} You are an expert image captioner. When describing the image, avoid these common errors: \texttt{\{hallucination\_notes\}}. Output only the caption.

\medskip
\textbf{User:} Describe this image in detail.
\end{tcolorbox}

\noindent\textbf{Stage 2: Detail-Focused Caption.}
\begin{tcolorbox}[promptbox]
\textbf{System:} You are an expert image captioner. Describe the image focusing on the aspects listed below, which are commonly overlooked. Only describe what is \textbf{CLEARLY VISIBLE} --- do not guess or infer. Output only the caption.

\medskip
\textbf{User:} Describe this image, paying special attention to these commonly missed aspects:\\[4pt]
\texttt{\{missing\_detail\_notes\}}
\end{tcolorbox}

\noindent\textbf{Stage 3: Merging Distinct Captions.}
\begin{tcolorbox}[promptbox]
\textbf{System:} You supplement a base caption with new information from a second caption.

\smallskip
Rules:
\begin{itemize}[leftmargin=1.8em, topsep=2pt, itemsep=1pt, parsep=0pt]
  \item The base caption is the foundation --- preserve its wording, counts, colors, and positions as-is.
  \item From the second caption, \textbf{only add} objects or elements \textbf{NOT already mentioned} in the base.
  \item Do \textbf{NOT} change any existing descriptions (counts, colors, spatial terms, materials).
  \item Verify each new element against the image before adding it.
  \item If the second caption has no genuinely new elements, return the base caption unchanged.
\end{itemize}

\medskip
\textbf{User:}\\
\texttt{Base caption: \{Grounded Base Caption\}}\\
\texttt{Second caption: \{Detail-Focused Caption\}}\\[4pt]
Add only new, verified elements from the second caption into the base. Do not modify existing details. Output only the final caption:
\end{tcolorbox}

\section{Inference Cost Details}
\label{sec:tflops_}
Beyond caption quality, practical deployment also requires efficient inference.
To evaluate this aspect, we measure cost-efficiency across different methods and open-source models in terms of TFLOPs per image.
Following Kaplan et al.~\cite{kaplan2020scaling}, we approximate the inference cost of a single forward pass as $C \approx 2NT$, where $N$ denotes the number of non-embedding parameters and $T$ is the total number of processed tokens.
For methods that involve multiple calls per image, image tokens are counted only once through KV caching.
As shown in Table~\ref{tab:opensource_cost_tflops}, ReflectCAP consistently achieves lower inference cost than CapMAS across all evaluated open-source models, while maintaining stronger captioning performance.
In particular, ReflectCAP requires 21--36\% less compute than CapMAS.

\begin{table}[htbp]
\centering
\small
\renewcommand{\arraystretch}{1.08}
\setlength{\tabcolsep}{8pt}
\caption{Compute cost of different methods on open-source models, measured in TFLOPs per final caption with KV caching.}
\label{tab:opensource_cost_tflops}
\begin{adjustbox}{width=0.5\columnwidth}
\begin{tabular}{llc}
\toprule
Model & Method & TFLOPs/final caption \\
\midrule
\multirow{3}{*}{Qwen3-VL-32B} 
& Zero-shot   & 219.97 \\
& CapMAS      & 490.83 \\
& ReflectCAP  & 356.48 \\
\midrule
\multirow{3}{*}{Qwen3-VL-8B} 
& Zero-shot   & 53.22 \\
& CapMAS      & 105.36 \\
& ReflectCAP  & 67.60 \\
\midrule
\multirow{3}{*}{Qwen2.5-VL-32B} 
& Zero-shot   & 280.00 \\
& CapMAS      & 533.68 \\
& ReflectCAP  & 385.41 \\
\midrule
\multirow{3}{*}{Qwen2.5-VL-7B} 
& Zero-shot   & 64.04 \\
& CapMAS      & 96.41 \\
& ReflectCAP  & 76.06 \\
\midrule
\multirow{3}{*}{InternVL3.5-38B} 
& Zero-shot   & 213.71 \\
& CapMAS      & 367.68 \\
& ReflectCAP  & 297.31 \\
\midrule
\multirow{3}{*}{InternVL3.5-4B} 
& Zero-shot   & 22.37 \\
& CapMAS      & 36.90 \\
& ReflectCAP  & 27.51 \\
\bottomrule
\end{tabular}
\end{adjustbox}
\end{table}

\section{Effect of Note Generator}
\label{appendix:note-author}

We investigate whether the quality of \textit{Structured Reflection Notes} improves when a more capable model serves as the note author.
We compare two conditions:
(1)~\textbf{Self-generated}, where the same model serves as the captioning agent, feedback agent, and note organizer---i.e., the target model analyzes its own errors and writes the error note itself;
and (2)~\textbf{GPT-4.1 mini (proxy writer)}, where the captioning agent remains the target model but GPT-4.1 mini replaces both the feedback agent and the note organizer, analyzing the target model's error patterns and writing the note on its behalf.

\begin{table}[t]
\centering
\caption{Self-generated error notes vs.\ notes written by GPT-4.1 mini as a proxy writer. Both note types describe the target model's error patterns; only the authorship differs. Bold indicates the better score per target model.}
\label{tab:note-author}
\begin{tabular}{llcccc}
\toprule
\textbf{Target Model} & \textbf{Note Source} & \textbf{Fact.} & \textbf{Cov.} & \textbf{F1} & \textbf{$\Delta$F1} \\
\midrule
\multirow{2}{*}{Qwen3-VL-8B}
  & Self-generated  & \textbf{77.6} & 72.4 & \textbf{74.9} & --- \\
  & GPT-4.1 mini    & 76.7 & \textbf{73.3} & \textbf{74.9} & $\pm$0.0 \\
\midrule
\multirow{2}{*}{Qwen3-VL-32B}
  & Self-generated  & 77.9 & \textbf{75.7} & 76.8 & --- \\
  & GPT-4.1 mini    & \textbf{81.0} & 75.4 & \textbf{78.1} & +1.3 \\
\midrule
\multirow{2}{*}{Qwen2.5-VL-7B}
  & Self-generated  & \textbf{68.1} & 63.5 & 65.7 & --- \\
  & GPT-4.1 mini    & 67.4 & \textbf{64.8} & \textbf{66.0} & +0.3 \\
\midrule
\multirow{2}{*}{Qwen2.5-VL-32B}
  & Self-generated  & \textbf{73.9} & 68.6 & 71.2 & --- \\
  & GPT-4.1 mini    & 73.0 & \textbf{70.6} & \textbf{71.8} & +0.6 \\
\midrule
\multirow{2}{*}{InternVL3.5-4B}
  & Self-generated  & \textbf{66.6} & \textbf{62.4} & \textbf{64.4} & --- \\
  & GPT-4.1 mini    & 65.2 & 62.2 & 63.7 & $-$0.7 \\
\midrule
\multirow{2}{*}{InternVL3.5-38B}
  & Self-generated  & \textbf{75.7} & 66.2 & \textbf{70.6} & --- \\
  & GPT-4.1 mini    & 72.7 & \textbf{66.8} & 69.6 & $-$1.0 \\
\bottomrule
\end{tabular}
\end{table}

Table~\ref{tab:note-author} shows that replacing the note author with a more capable model does not uniformly improve performance. While Qwen3-VL-32B sees a notable gain (+1.3 F1), the remaining configurations show only marginal improvements or even slight degradation, particularly in the InternVL3.5 family.

A qualitative comparison of the generated notes offers a possible explanation.
GPT-4.1 mini tends to produce general principle-level guidelines such as ``\textit{Avoid speculative or inferred details about materials, styles, or dates},'' whereas self-generated notes are more case-specific, e.g., ``\textit{Do not exaggerate water clarity or infer bottom composition (e.g., `sandy/silty').}''
This suggests that each model may respond better to a particular note style, and that a universally stronger author does not guarantee a better-fitting note.

\begin{figure}[t]
\centering
\begin{minipage}[c]{0.45\columnwidth}
  \centering
  \includegraphics[width=\linewidth]{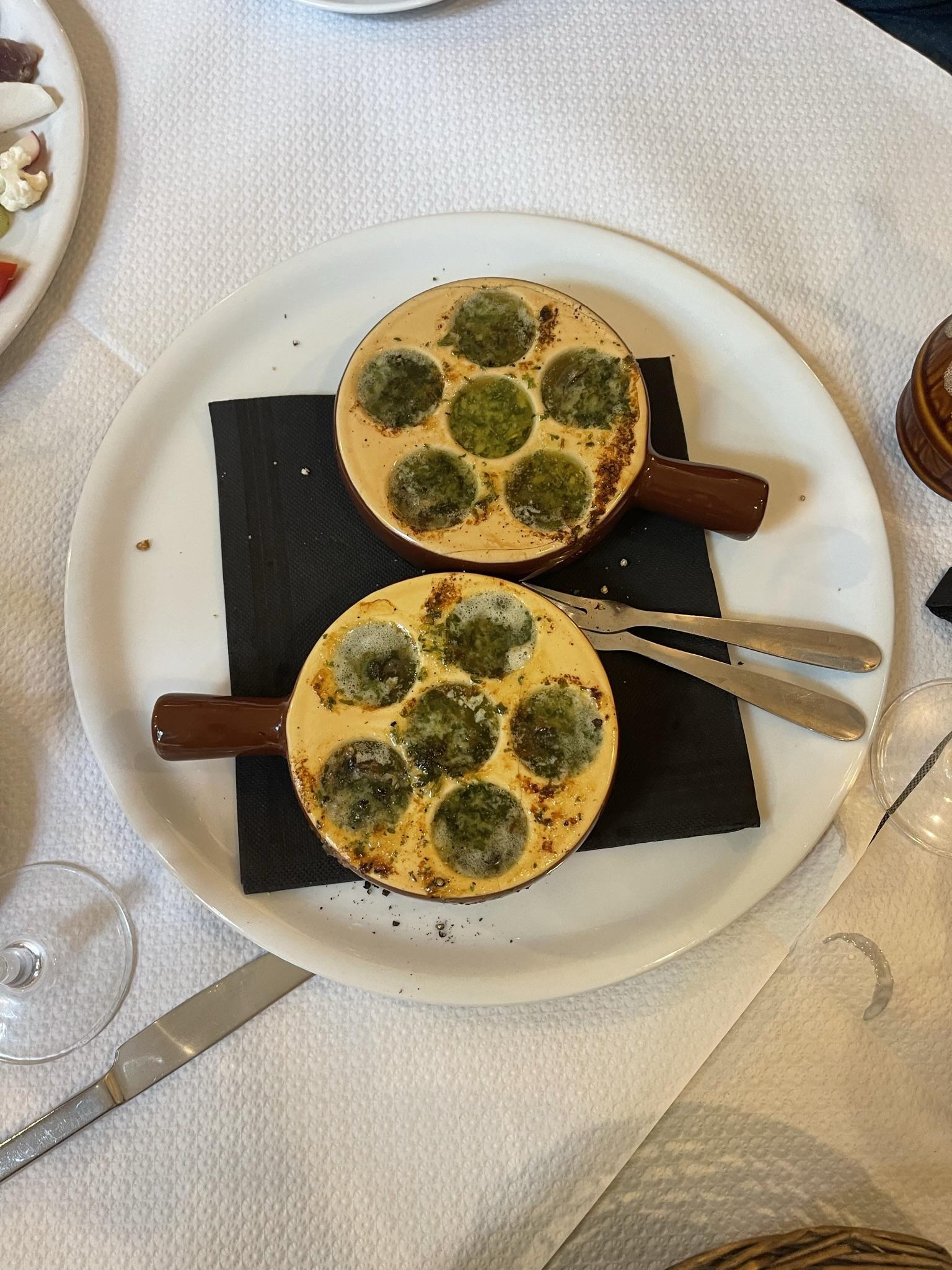}
\end{minipage}%
\hfill
\begin{minipage}[c]{0.5\columnwidth}
  \footnotesize
  \textbf{Zero-shot caption}  \\[1pt]
  \textcolor{red}{\ding{55}}~``\textit{each dish contains \underline{nine} circular indentations}'' \\
  \textcolor{red}{\ding{55}}~``\textit{a soft, \underline{custard-like} substance topped with green herb garnish}'' \\[3pt]
  \textbf{Grounded base caption}  \\[1pt]
  \textcolor{green!60!black}{\ding{51}}~``\textit{baked escargot}'' {\scriptsize (correct food identification)} \\
  \textcolor{green!60!black}{\ding{51}}~{\scriptsize No count claim --- avoids fabrication} \\[3pt]
  \textbf{ReflectCAP}  \\[1pt]
  \textcolor{blue}{\textbf{+}}~``\textit{Shadows from overhead lighting fall across the table}''~\textcolor{green!60!black}{\ding{51}} \\
  \textcolor{blue}{\textbf{+}}~``\textit{A fork and knife rest beside the lower ramekin}''~\textcolor{green!60!black}{\ding{51}} \\[1pt]
  \rule{\linewidth}{0.4pt} \\[2pt]
  {\scriptsize \textbf{Error notes triggered:} \textit{Mention visible lighting positions}; \textit{Confirm exact placement via spatial anchors} --- both within the target model's capability.}
\end{minipage}
\caption{Success case. Error notes correct zero-shot hallucinations (\textcolor{red}{red}~$\to$~\textcolor{green!60!black}{green}), and the extract-merge step successfully adds verifiable details (\textcolor{blue}{blue},~\textcolor{green!60!black}{\ding{51}}).}
\label{fig:case_success}
\end{figure}

\begin{figure}[htbp]
\centering
\begin{minipage}[c]{0.45\columnwidth}
  \centering
  \includegraphics[width=\linewidth]{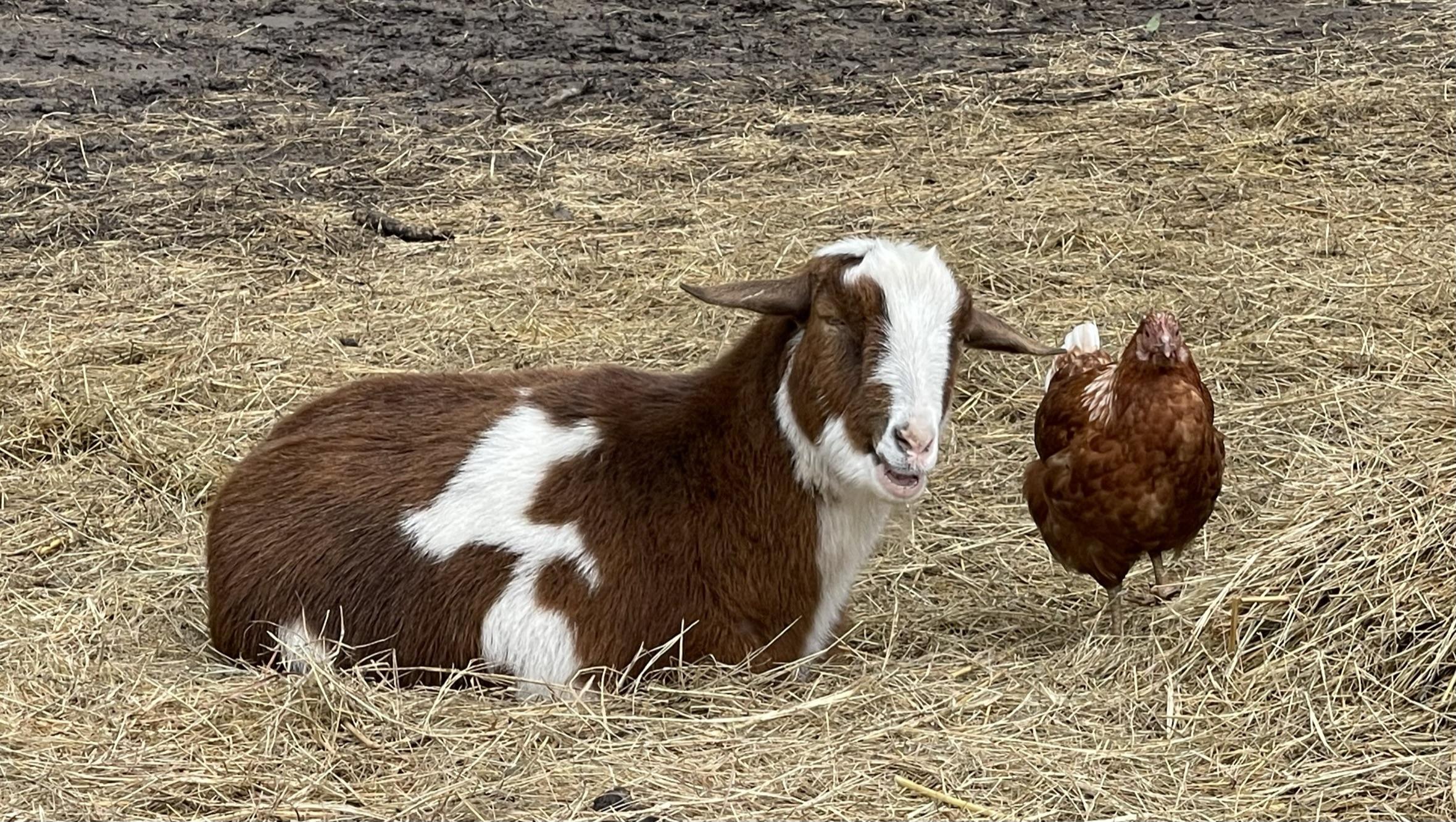}
\end{minipage}%
\hfill
\begin{minipage}[c]{0.5\columnwidth}
  \footnotesize
  \textbf{Zero-shot caption} \\[1pt]
  \textcolor{red}{\ding{55}}~``\textit{a prominent \underline{star-shaped} patch across its back}'' \\
  \textcolor{red}{\ding{55}}~``\textit{body facing \underline{away} from the camera \dots head turned to look back over its shoulder}'' \\[3pt]
  \textbf{Grounded base caption}  \\[1pt]
  \textcolor{green!60!black}{\ding{51}}~``\textit{a large white patch on its side}'' \\
  \textcolor{green!60!black}{\ding{51}}~``\textit{standing nearby to the right}'' \\[3pt]
  \textbf{ReflectCAP}  \\[1pt]
  \textcolor{blue}{\textbf{+}}~``\textit{The chicken is positioned slightly \underline{higher} than the goat's head level}''~\textcolor{red}{\ding{55}} \\[1pt]
  \rule{\linewidth}{0.4pt} \\[2pt]
  {\scriptsize \textbf{Error note triggered:} \textit{Confirm exact placement via spatial anchors} --- beyond the target model's reliable capability.}
\end{minipage}
\caption{Limitation case. Error notes correct zero-shot hallucinations (\textcolor{red}{red}~$\to$~\textcolor{green!60!black}{green}), but the extract-merge step introduces a new spatial error (\textcolor{blue}{blue},~\textcolor{red}{\ding{55}}) when following a missing-detail note that exceeds the target model's perceptual competence.}
\label{fig:case_limitation}
\end{figure}

\section{Qualitative Analysis}
Figures~\ref{fig:case_success} and~\ref{fig:case_limitation} present two contrasting examples on Qwen3-VL-8B that together illustrate the strengths and limitations of ReflectCAP.
In both cases, error notes successfully suppress zero-shot hallucinations: the grounded base caption corrects a fabricated indentation count and a food misidentification in Figure~\ref{fig:case_success}, and removes an invented coat pattern and a reversed body orientation in Figure~\ref{fig:case_limitation}.
The difference emerges when merging the detail caption into the base caption.
In Figure~\ref{fig:case_success}, the note prompts the model to describe overhead lighting and silverware placement, and the model accurately incorporates these details, improving coverage without sacrificing factuality.
In contrast, Figure~\ref{fig:case_limitation} shows that the same type of note (e.g., ``\textit{Confirm exact placement via spatial anchors}'') instead leads to a fabricated height comparison between the chicken and the goat.
This illustrates that reflection notes can guide the model to attend to previously overlooked details, but whether this results in faithful descriptions or additional hallucinations depends on the model's perceptual ability.

Currently, verifying the factuality of newly added details is left to the model itself during the merging step, which proves insufficient when the model lacks the visual understanding ability to accurately perceive the prompted content.
A more explicit verification mechanism at this stage could make ReflectCAP a more robust framework that effectively boosts both coverage and factuality.

\section{Domain-Specific Captioning with ReflectCAP}
\label{sec:appendix_domain}

The main experiments focus on detailed captioning of everyday images, but ReflectCAP's Structured Reflection Notes are not tied to everyday image captioning---they adapt automatically to the exemplar set provided in the offline phase. To verify this, we apply ReflectCAP to fashion product captioning, which differs substantially from everyday image captioning in both visual characteristics and description conventions.

Specifically, we use Fashion-Gen~\cite{rostamzadeh2018fashiongengenerativefashiondataset}, a
large-scale dataset of 293,008 high-resolution studio fashion images paired
with paragraph-level captions authored by professional stylists covering
fine-grained garment attributes such as fabric, cut, fit, closures, and color.
Unlike everyday captions that freely describe scenes, spatial layouts, and
background context, fashion captions focus on the design specification of a
single item, making the domain shift explicit. We select 30 exemplar images
from Fashion-Gen for the offline phase and analyze how the resulting notes
differ from those constructed on everyday images.

\noindent\textbf{Comparison of Structured Reflection Notes.} Table~\ref{tab:domain_notes} presents the Avoid and Include notes generated by
the same pipeline (GPT-4.1-mini as the target LVLM) on everyday images versus
fashion images. Without any modification to the framework, the notes shift from
scene-level guidance (\emph{e.g.}, ``Avoid inferring lighting direction or time
of day'') to garment-level guidance (\emph{e.g.}, ``Do not add clothing fit or
garment length details not clearly visible''). Notably, the fashion Include
notes capture domain-specific conventions that have no counterpart in everyday
captioning, such as interior finishing details (\emph{e.g.}, lining, surgeon's
cuffs) and precise pattern or fabric texture names. This demonstrates that the offline phase automatically distills domain-adapted Structured Reflection Notes from a small exemplar set—once 30 images with domain-specific reference captions are provided, no further manual prompt engineering or domain expertise is required.

\begin{table*}[htbp]
\centering
\caption{\textbf{Structured Reflection Notes: Scene-level vs.\ Product-level captioning.}
Both note sets are generated by the same offline pipeline with GPT-4.1-mini.
The notes automatically adapt to domain-specific visual characteristics and
description conventions.}
\label{tab:domain_notes}
\small
\renewcommand{\arraystretch}{1.25}
\newcommand{\grayrow}{\noalign{\vskip 2pt{\color{gray!35}\hrule}\vskip 4pt}}
\begin{tabular}{@{}p{0.44\textwidth}@{\hspace{4pt}}!{\color{gray!50}\vrule}@{\hspace{8pt}}p{0.44\textwidth}@{}}
\toprule
\textbf{Scene Level Captioning (DOCCI)} & \textbf{Product Level Captioning (Fashion-Gen)} \\
\midrule
\multicolumn{2}{@{}l}{\textit{Avoid Notes} $\mathcal{N}_{\text{avoid}}$} \\
\midrule
Avoid specifying materials, object types, or inferred object roles without clear visible evidence. &
Match exact color, pattern, and fabric details precisely from reference or visible image. \\
\grayrow
Do not add or alter visible object features, colors, or counts without confirmation. &
Do not add clothing fit, style, or garment length details not explicitly stated or clearly visible. \\
\grayrow
Avoid inferring lighting direction, time of day, or environment unless clearly visible. &
Avoid subjective or interpretive descriptions not supported by reference or image. \\
\grayrow
Do not add unsupported details to signs, logos, or symbols. &
Do not add accessories, footwear, or personal attributes unless clearly visible or mentioned. \\
\grayrow
Avoid assumptions about scene context or climate not explicitly shown. &
Avoid describing visible details that contradict the reference (\emph{e.g.}, visible buttons when concealed). \\
\midrule
\multicolumn{2}{@{}l}{\textit{Include Notes} $\mathcal{N}_{\text{include}}$} \\
\midrule
Include precise object condition, orientation, posture, and spatial positioning as visible. &
Include all key garment design features and construction details (\emph{e.g.}, collars, closures, pockets, vents, cuffs). \\
\grayrow
Describe all visible background elements, environmental context, and relevant landscape features. &
Specify distinctive branding, logos, patches, or signature elements described in the reference. \\
\grayrow
Note lighting effects, shadows, reflections, and surface details accurately. &
Mention precise color, pattern names, fabric texture, tonal stitching, and finishing details exactly as given. \\
\grayrow
Mention detailed features of signs, logos, fonts, and text including illegible or small elements. &
Include garment fit and silhouette terms exactly as described (\emph{e.g.}, shift dress, slim-fit trousers). \\
\grayrow
Include counts, distribution, distinctive markings, and compositional or stylistic shot details. &
Add all notable interior and finishing details (\emph{e.g.}, lining, buttonholes, surgeon's cuffs, belt details). \\
\bottomrule
\end{tabular}
\end{table*}

\begin{figure*}[htbp]
\centering
 
\fbox{\begin{minipage}[t]{0.99\textwidth}
\vspace{4pt}
\begin{minipage}[t]{0.25\textwidth}
\vspace{0pt}
\centering
\includegraphics[width=\linewidth]{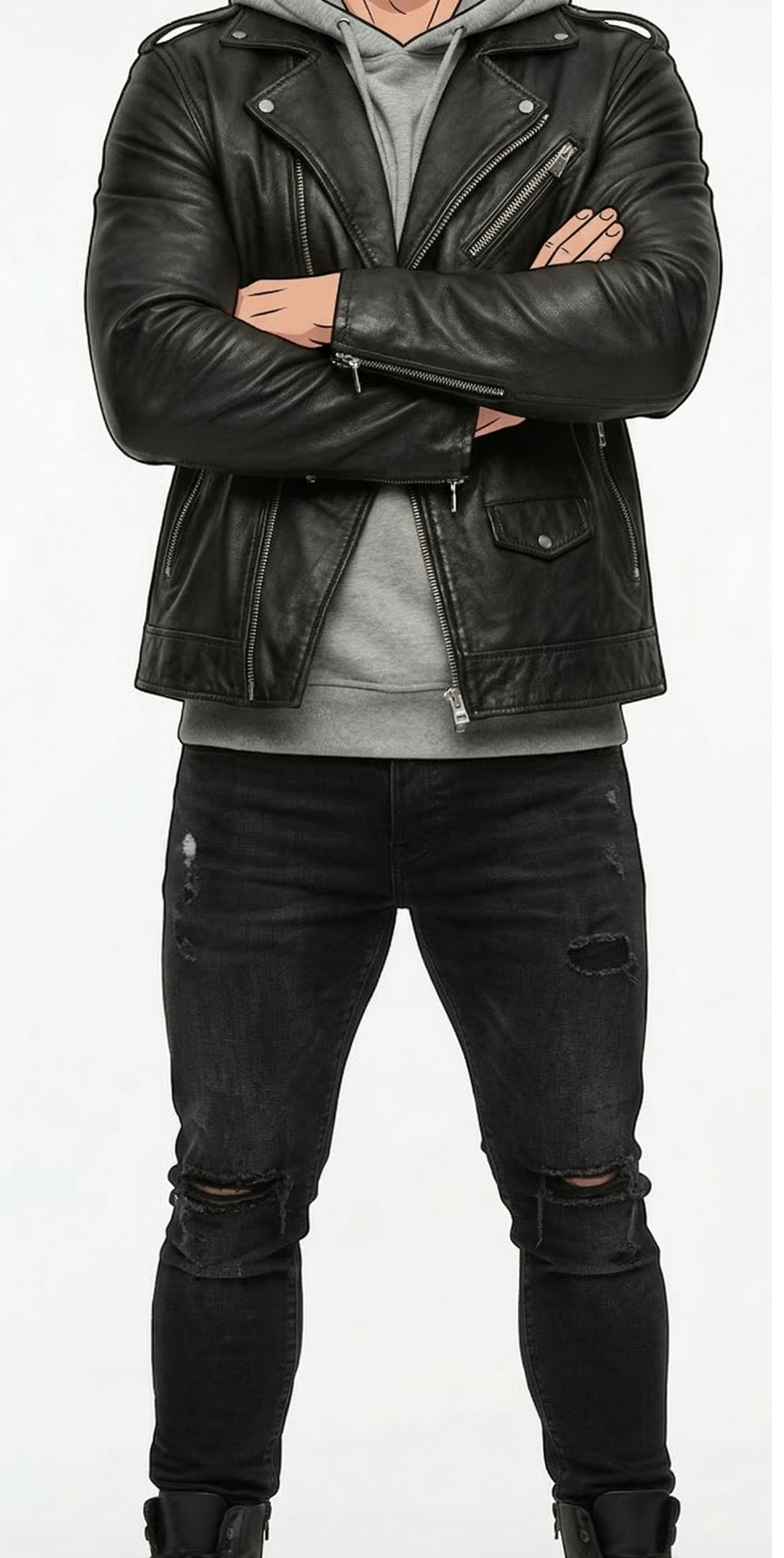}
\end{minipage}%
\hfill
\begin{minipage}[t]{0.7\textwidth}
\vspace{0pt}
 
\textbf{1) Zero-shot Caption}\\[1pt]
{\scriptsize ``The image shows a person standing with their arms crossed. They are wearing a black leather jacket over a gray hooded sweatshirt. The jacket has multiple zippers and buttons, giving it a rugged look. The person is also wearing black, ripped skinny jeans and black shoes.''}
 
\vspace{3pt}
\textbf{2) ReflectCAP Caption}\\[1pt]
{\scriptsize ``The image shows a person standing with their arms crossed, dressed in a black leather biker jacket with a classic \textcolor{recgreen}{asymmetrical front zipper closure} over a grey hooded sweatshirt with a visible hood and drawstrings. The jacket features \textcolor{recgreen}{silver zippers}, a \textcolor{recgreen}{wide lapel collar with snap-button details on the collar points}, \textcolor{recgreen}{shoulder epaulets with snap buttons}, \textcolor{recgreen}{zippered cuffs}, \textcolor{recgreen}{tonal stitching throughout}, and a \textcolor{recgreen}{flap pocket with a snap button}. There are multiple zippered pockets: one \textcolor{recgreen}{angled chest pocket on the left side} and two \textcolor{recgreen}{vertical zippered pockets near the waist} on each side. The person is also wearing black, distressed skinny jeans with rips at the knees and some frayed areas on the thighs. The jeans taper down to the ankles, where they meet \textcolor{recgreen}{black lace-up boots with a matte finish}.''}
\end{minipage}
 
\vspace{4pt}
{\footnotesize
\begin{tabular}{@{\hspace{6pt}}p{0.93\textwidth}@{}}
\textit{Reflection Notes Applied} \\
\hline
``Include all key garment design features and construction details (collars, closures, pockets, cuffs)'' \\
``Mention precise color, fabric texture, tonal stitching, and finishing details'' \\
\end{tabular}
}
\vspace{2pt}
\end{minipage}}
 
\vspace{5pt}
 
\fbox{\begin{minipage}[t]{0.99\textwidth}
\vspace{4pt}
\begin{minipage}[t]{0.25\textwidth}
\vspace{0pt}
\centering
\includegraphics[width=\linewidth]{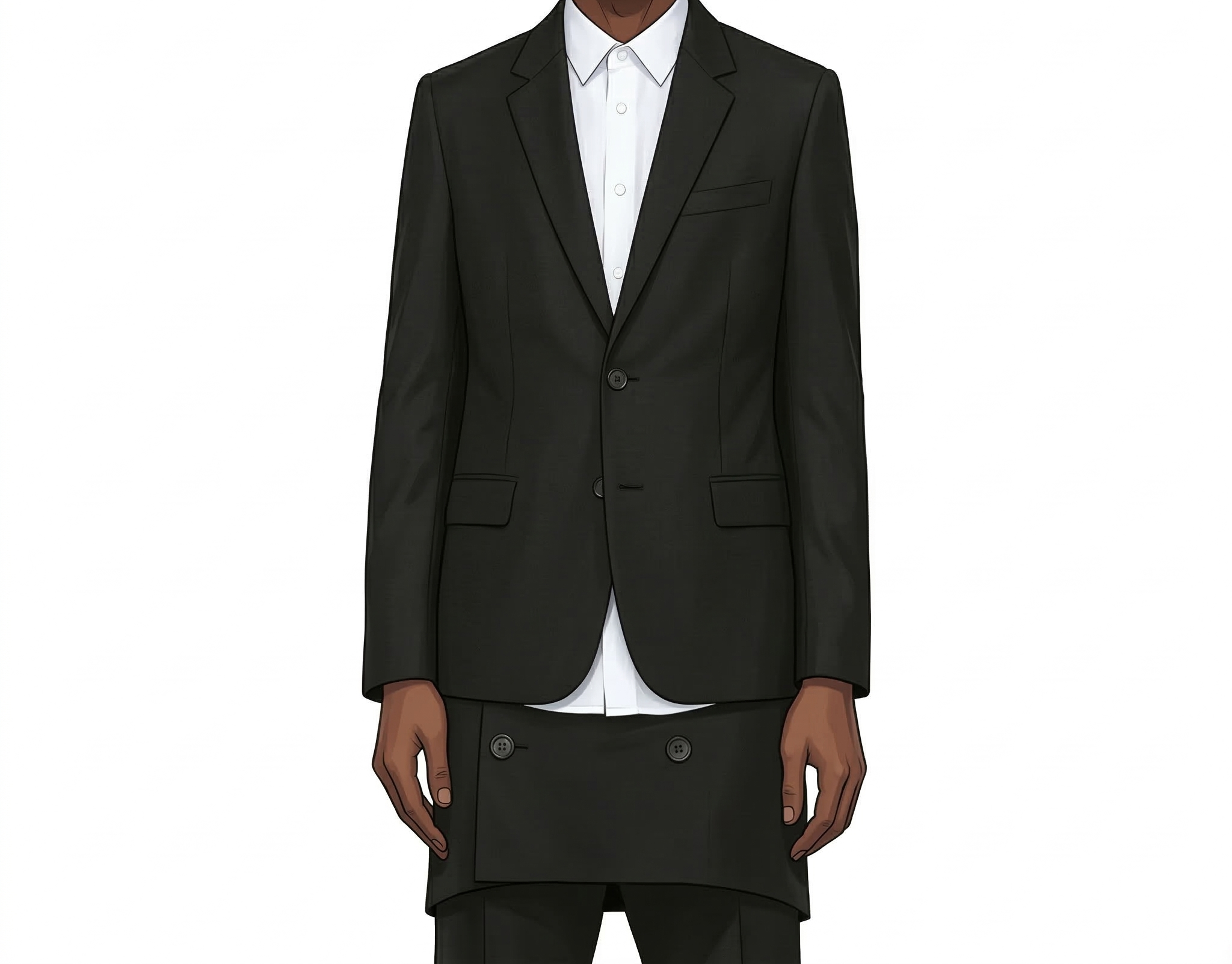}
\end{minipage}%
\hfill
\begin{minipage}[t]{0.7\textwidth}
\vspace{0pt}
 
\textbf{1) Zero-shot Caption}\\[1pt]
{\scriptsize ``The image shows a person wearing a tailored black suit jacket over a white button-up shirt. The jacket has a classic lapel, two front buttons (of which the top one is buttoned), and flap pockets on each side. Underneath the jacket, the person is wearing an additional black garment that resembles a wrap-around skirt or apron with two visible buttons in the front.''}
 
\vspace{3pt}
\textbf{2) ReflectCAP Caption}\\[1pt]
{\scriptsize ``The image depicts a person wearing a formal outfit consisting of a black tailored blazer with a classic \textcolor{recgreen}{notch lapel collar} and a \textcolor{recgreen}{two-button closure} over a white dress shirt. The blazer features \textcolor{recgreen}{flap pockets on both sides}, a \textcolor{recgreen}{welt chest pocket on the left}, and long sleeves with no visible buttons or cuff details. The blazer has a \textcolor{recgreen}{smooth, matte finish with tonal stitching} that blends seamlessly with the fabric, providing a \textcolor{recgreen}{tailored but not overly slim, structured silhouette}. Underneath the blazer, the white dress shirt has a \textcolor{recgreen}{pointed collar} and visible white buttons down the front placket, buttoned up to the neck. The bottom part of the outfit includes a unique black garment resembling a wrap or apron with \textcolor{recgreen}{two large black buttons} at the front, creating an asymmetrical hemline. The fabric appears to be a \textcolor{recgreen}{smooth woven material, likely wool or a wool blend}, with no visible texture, pattern, branding, logos, patches, or signature elements.''}
\end{minipage}
 
\vspace{4pt}
{\footnotesize
\begin{tabular}{@{\hspace{6pt}}p{0.93\textwidth}@{}}
\textit{Reflection Notes Applied} \\
\hline
``Include all key garment design features and construction details (collars, closures, pockets)'' \\
``Do not add clothing fit, style, or garment length details not explicitly stated or clearly visible'' \\
``Mention precise fabric texture, tonal stitching, and finishing details'' \\
\end{tabular}
}
\vspace{2pt}
\end{minipage}}
 
\vspace{2pt}
\caption{\textbf{Fashion domain qualitative examples.}
Zero-shot captions produce generic descriptions (\emph{e.g.}, ``multiple zippers and buttons,'' ``a classic lapel''),
while ReflectCAP generates domain-appropriate captions with precise garment construction vocabulary.
\textcolor{recgreen}{Green} denotes fashion-specific details recovered by the Structured Reflection Notes.}
\label{fig:fashion_case}
\end{figure*}

\noindent\textbf{Qualitative Examples.}
Figure~\ref{fig:fashion_case} presents two representative examples comparing
zero-shot and ReflectCAP captions on AI-generated fashion illustrations,
using the Structured Reflection Notes learned from real Fashion-Gen
photographs. In both cases, the
zero-shot baseline produces generic visual descriptions---\emph{e.g.}, ``black
leather jacket with multiple zippers and buttons'' or ``tailored black suit
jacket with a classic lapel.'' In contrast, ReflectCAP
generates domain-appropriate captions with garment construction vocabulary
(\emph{e.g.}, asymmetrical front zipper closure, notch lapel, welt chest pocket, shoulder epaulets), precise material
descriptions (\emph{e.g.}, silver zippers, snap-button details, tonal stitching), and accurate
fit terminology (\emph{e.g.}, structured silhouette, skinny jeans that taper to the ankles). These
domain-specific details are elicited not by manual prompt engineering but by the
automatically generated Structured Reflection Notes, which direct the model to
attend to garment construction features and suppress unsupported fit
descriptions. This confirms that ReflectCAP's note-guided approach generalizes
beyond everyday image captioning to specialized visual domains.

\section{Discussion}
\noindent\textbf{Perceptual Boundary and Verification.}
The Include Notes in ReflectCAP guide the model to describe details it typically overlooks, but when this guidance exceeds the model's visual perception capability, it can instead introduce new hallucinations. The current merging step adopts a conservative strategy that prioritizes the base caption, yet it has limitations in fully filtering out hallucinations introduced from the detail caption. Moreover, when the two captions conflict, the framework is designed to trust the base caption, but the base caption itself is not guaranteed to be always accurate, allowing incorrect descriptions to persist in the final output. If an external verifier or visual grounding module were introduced at the merging stage to independently verify details from both captions, it would become possible to aggressively expand coverage while preserving factuality, pushing the current factuality--coverage Pareto frontier further.

\noindent\textbf{Domain-Specific Captioning.}
As demonstrated in Appendix~\ref{sec:appendix_domain}, ReflectCAP's Structured Reflection Notes adapt to the fashion domain simply by replacing the exemplar set, without any modification to the framework. However, the current experiment is limited to qualitative analysis on a single domain, lacking quantitative evaluation. If validated across multiple domains such as medical imaging, remote sensing, and e-commerce with domain-specific evaluation protocols, ReflectCAP could establish itself as a general-purpose framework that can be immediately deployed to diverse specialized domains with only a small set of exemplars, without training dedicated captioning models for each domain.

\noindent\textbf{Training Data Generation.}
As shown in Section 5.1, fine-tuning with ReflectCAP-generated captions maintains factuality while improving coverage compared to human-authored captions. This is because ReflectCAP generates captions within the model's perceptual boundary, avoiding forcing details the model cannot actually perceive and thus suppressing hallucination amplification. Scaling this property, ReflectCAP can be extended into a pipeline for generating high-quality caption data for T2I/T2V training without human annotation. In particular, when combined with the domain-specific adaptation discussed above, this could simultaneously address the scarcity of training data in specialized domains.

\end{document}